\documentclass[10pt,twocolumn,letterpaper]{article}

\usepackage{cvpr}              

\definecolor{cvprblue}{rgb}{0.21,0.49,0.74}
\usepackage[pagebackref,breaklinks,colorlinks,allcolors=cvprblue]{hyperref}

\usepackage{svg} 
\usepackage{url}
\usepackage{booktabs}
\usepackage{multirow}
\usepackage{textcomp}
\usepackage{graphicx}
\usepackage{amssymb}
\usepackage{float}
\usepackage{bm}
\usepackage{xcolor}
\usepackage{hyperref}
\usepackage[accsupp]{axessibility}  

\usepackage{amsthm}
\newtheorem{theorem}{Theorem}
\usepackage{algorithmic}
\usepackage{algorithm}
\usepackage{bbm}

\def\O{\bm{ \mathrm O}}

\def\a{\mathbf{a}}
\def\b{\mathbf{b}}

\def\med{{ \mathrm {median}}}
\def\agg{{ \mathrm {DSF}}}
\def\mean{{ \mathrm {mean}}}
\def\smoothMax{{ \mathrm {max}}}
\def\smoothArgMax{{ \mathrm {amax}}}
\def\smoothMin{{ \mathrm {min}}}
\def\smoothMax{{ \mathrm {max}}}
\def\smoothArgMin{{ \mathrm {amin}}}
\def\smoothArgMed{{ \mathrm {amedian}}}

\def\paperID{9462} 
\def\confName{CVPR}
\def\confYear{2025}

\title{Semi-Supervised State-Space Model with Dynamic Stacking Filter for Real-World Video Deraining}

\author{\textbf{Shangquan Sun}$^{1,2}$\hspace{7mm}
\textbf{Wenqi Ren}$^{3,4,5}$\hspace{7mm}
\textbf{Juxiang Zhou}$^{6}$ \\
\hspace{6.7mm}\textbf{Shu Wang}$^{7}$\hspace{11.5mm}
\textbf{Jianhou Gan}$^{6}$\hspace{8.3mm}
\textbf{Xiaochun Cao}$^{3\dag}$ \\
$^{1}$Institute of Information Engineering, Chinese Academy of Sciences, Beijing, China \\ 
$^{2}$School of Cyber Security, University of Chinese Academy of Sciences, Beijing, China \\ 
$^{3}$School of Cyber Science and Technology, Shenzhen Campus of Sun Yat-sen University \\
$^{4}$MoE Key Laboratory of Information Technology \\
$^{5}$Guangdong Provincial Key Laboratory of Information Security Technology \\
$^{6}$Key Laboratory of Educational Information for Nationalities, Yunnan Normal University\\
$^{7}$School of Mechanical Engineering and Automation, Fuzhou University\\
{\tt\small shangquansun@gmail.com, }{\tt\small \{renwq3,caoxiaochun\}@mail.sysu.edu.cn}
}

\begin{document}
\maketitle
\vspace{-5mm}
\renewcommand{\thefootnote}{}
\footnotetext{\dag Corresponding author.}

\begin{abstract}
Significant progress has been made in video restoration under rainy conditions over the past decade, largely propelled by advancements in deep learning. 
Nevertheless, existing methods that depend on paired data struggle to generalize effectively to real-world scenarios, primarily due to the disparity between synthetic and authentic rain effects. 
To address these limitations, we propose a dual-branch spatio-temporal state-space model to enhance rain streak removal in video sequences. 
Specifically, we design spatial and temporal state-space model layers to extract spatial features and incorporate temporal dependencies across frames, respectively. 
To improve multi-frame feature fusion, we derive a dynamic stacking filter, which adaptively approximates statistical filters for superior pixel-wise feature refinement. 
Moreover, we develop a median stacking loss to enable semi-supervised learning by generating pseudo-clean patches based on the sparsity prior of rain. 
To further explore the capacity of deraining models in supporting other vision-based tasks in rainy environments, we introduce a novel real-world benchmark focused on object detection and tracking in rainy conditions. 
Our method is extensively evaluated across multiple benchmarks containing numerous synthetic and real-world rainy videos, consistently demonstrating its superiority in quantitative metrics, visual quality, efficiency, and its utility for downstream tasks.

\end{abstract}


\vspace{-5mm}
\abovedisplayshortskip=2.7pt
\belowdisplayshortskip=2.7pt
\abovedisplayskip=2.7pt
\belowdisplayskip=2.7pt
\vspace{-1mm}
\section{Introduction}
\vspace{-2mm}

\begin{figure}[t]
  \centering
   \includegraphics[width=\linewidth]{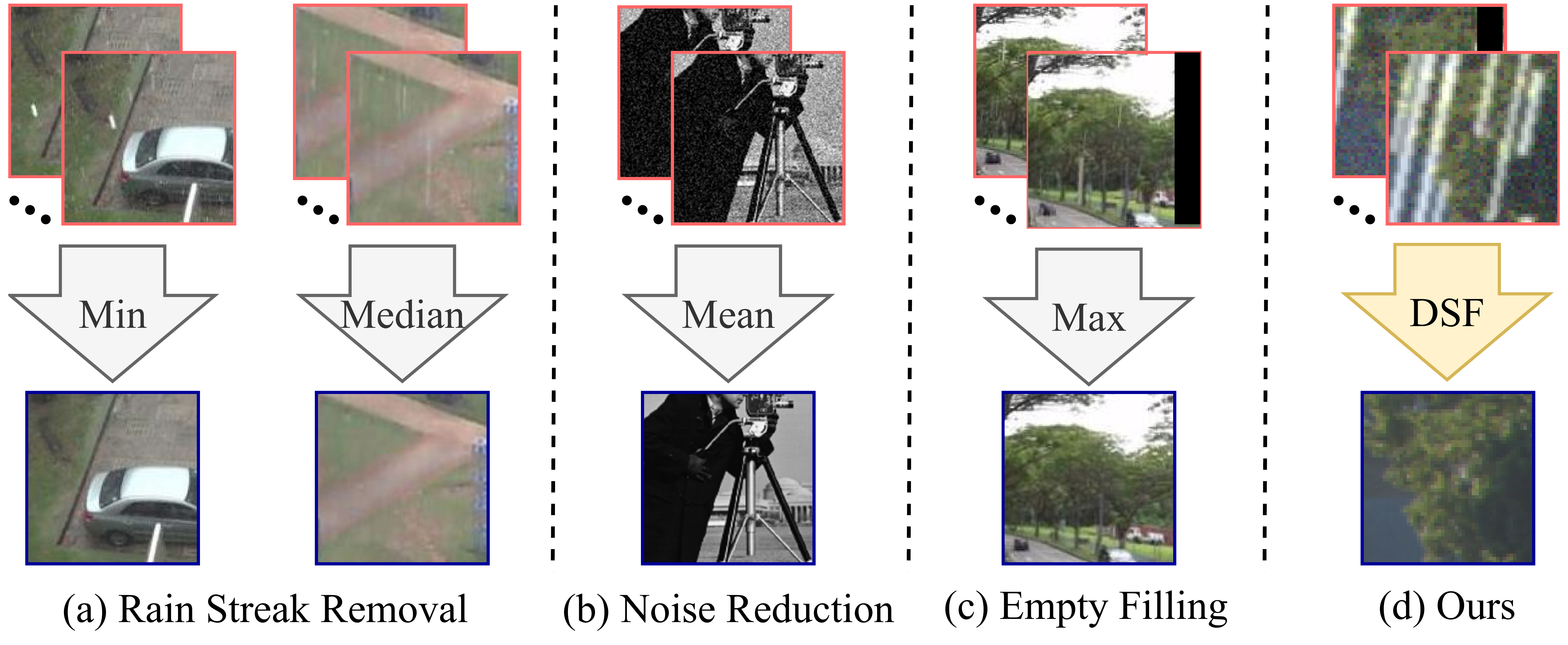}
\setlength{\abovecaptionskip}{-6pt} 
\vspace{-2mm}
   \caption{
   The functionality of various stacking filters in video restoration. 
   With aligned candidate frames, (a) shows that the min and median filters effectively remove falling rain streaks, while (b) and (c) illustrate how the mean and max filters aid in noise reduction and filling missing pixels, respectively. 
   Our dynamic stacking filter can adaptively approximate these filters at the pixel level, facilitating versatile frame fusion.
   }
   \label{fig:stacking}
   \vspace{-4mm}
\end{figure}

\looseness-1
Rain streaks in captured videos or images obscure the background, diminishing image quality and potentially impairing downstream high-level vision tasks such as object detection~\cite{hnewa2020object, sindagi2020prior,sun2022rethinking} and tracking~\cite{hassaballah2020vehicle}. 
Given the prevalence of rain as an adverse weather condition, effective rain streak removal is essential.

Since the advent of deep learning, numerous architectures have been developed to tackle this issue, including convolutional neural networks (CNNs)~\cite{Chen2018RobustCNN, Yang2020Self, Yue2021Semi}, Transformers~\cite{sun2023event, Chen2023drsformer}, and sequential models~\cite{wu2024rainmamba}. 
Despite these advancements, two key challenges persist: the disparity between synthetic and real-world rain streaks, and the practical application of deraining methods in real-world downstream tasks.
%
Model-based methods can effectively remove rain streaks from real-world rainy frames by solving complex, prior-based objective functions~\cite{Jiang2017NovelTensorBased, Wei2017Should, Jiang2019FastDerain}.
However, the high computational demands and manually defined hyper-parameters of model-based methods limit their real-world applicability and generalization. 
Conversely, data-driven approaches that rely on paired synthetic data struggle to bridge the gap between synthetic and real-world rain streaks, thus failing to achieve effective rain removal in real-world scenarios~\cite{Yue2021Semi, wu2024rainmamba}. 
Additionally, despite the existence of several synthetic benchmarks~\cite{jiang2023dawn, sun2023event}, the potential of leveraging video deraining to enhance other vision tasks in real-world rainy conditions remains largely unexplored.

\begin{figure}[t]
  \centering
   \includegraphics[width=1\linewidth]{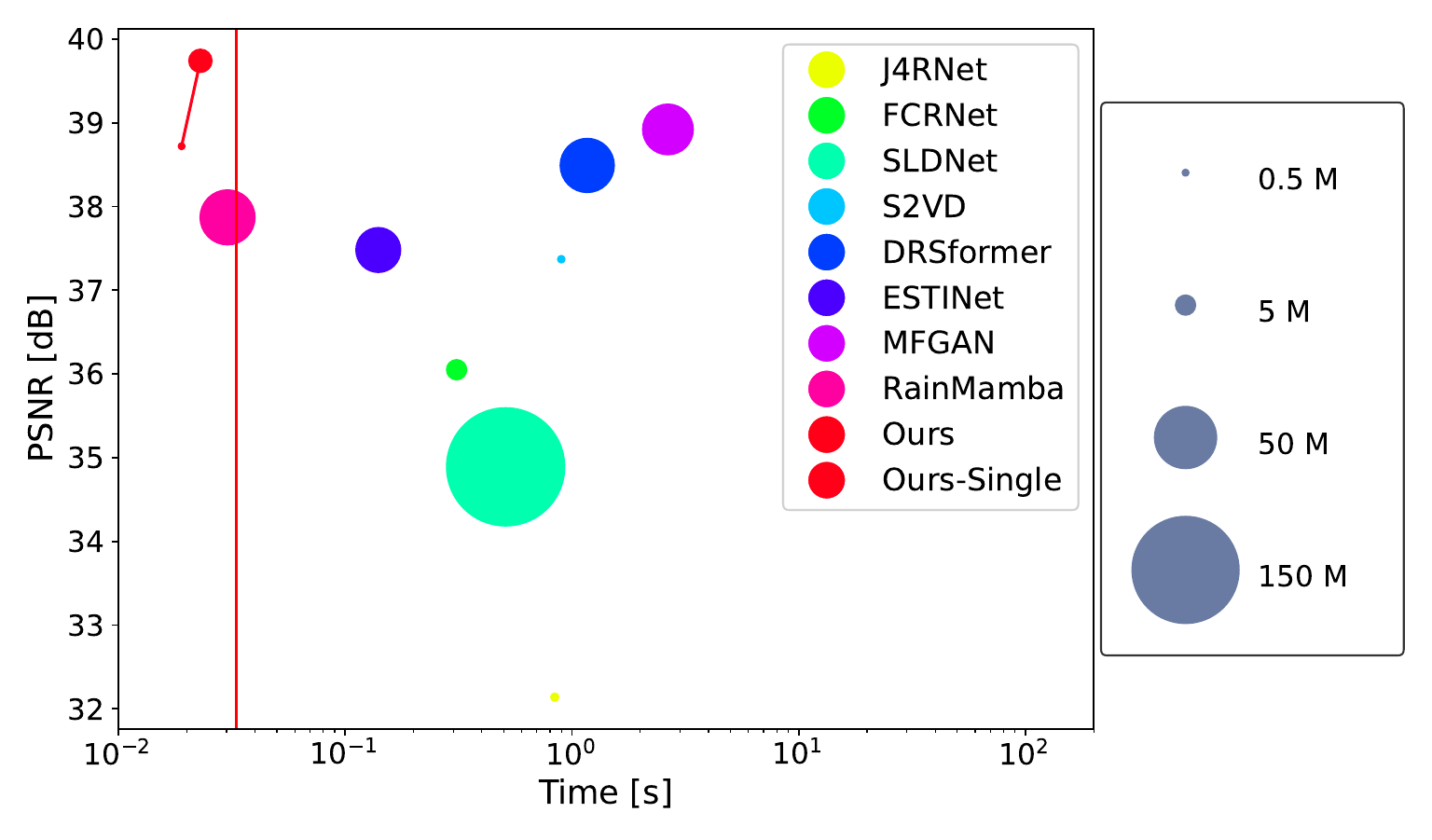}
\vspace{-8mm}
   \caption{The comparison of performance, time and space complexities among video deraining methods.
   The red line denotes the boundary of real-time inference.}
   \label{fig:complexity}
   \vspace{-3mm}
\end{figure}

To address these challenges, we introduce a semi-supervised dual-branch spatio-temporal state-space model for video deraining, named VDMamba.
Specifically, we implement one-step multi-frame warping by incorporating degradation-free optical flow knowledge into our model. 
Our dual-branch architecture comprises a spatial feature extraction branch and a temporal feature fusion branch, empowered by two key components: the spatial state-space model layer (S3ML) and the temporal state-space model layer (TSML). 
To further improve multi-frame information fusion, we derive a stacking-based dynamic filter that selectively integrates pixels from neighboring frames. 
This filter adaptively approximates various statistical functions, including the min and median filters for rain removal, the mean filter for noise suppression, and the max filter for filling misaligned pixels, as illustrated in Fig.~\ref{fig:stacking}. 
Utilizing the sparsity prior of rain, we incorporate a median stacking loss to enable semi-supervision, thereby enhancing generalization for effective rain streak removal in real-world scenarios.
As shown in Fig.~\ref{fig:complexity}, our method achieves state-of-the-art performance in video deraining with real-time inference speed and a compact model size.
To further assess the effectiveness of video deraining in supporting rain-affected downstream tasks, we establish a benchmark for object detection and tracking in rainy videos. 
%
%

\looseness-1
In summary, our contributions are as follows:
\begin{itemize}
    \item We introduce a spatio-temporal state-space model for video deraining, featuring two branches dedicated to spatial feature refinement and temporal information integration, respectively.
    
    \item To achieve improved real-world deraining, we develop a dynamic stacking filter for adaptive pixel aggregation, along with a semi-supervised stacking loss function based on the sparsity prior of rain.

    \item 
    We also create a real-world rainy video benchmark to assess the effectiveness of deraining methods in facilitating downstream tasks of object detection and tracking. 
    Extensive experiments show our method's superiority across synthetic, real-world, and downstream benchmarks.
\end{itemize}

\section{Related Works}

\noindent\textbf{Single Image Deraining.}
The rainy image is typically modeled as an additive composition~\cite{kang2012automatic,kim2013single},
\begin{equation}
    \mathbf{I} = \mathbf{B} + \mathbf{R},
\end{equation}
where $\mathbf{I}$, $\mathbf{B}$, and $\mathbf{R}\in \mathbb{R}^{H\times W\times 3}$ represent the rainy image, clean background, and rain streak layer, respectively. 
Due to the occlusion introduced by $\mathbf{R}$, recovering $\mathbf{B}$ becomes an ill-posed and inherently challenging task.
Numerous approaches have been proposed over the years, primarily falling into two categories: model-based and data-driven methods. 
Model-based techniques aim to decompose the rain and background layers, employing methods equipped with different priors~\cite{kang2012automatic,kim2013single,chen2014visual,luo2015removing}
In contrast, data-driven methods have recently emerged and largely dominated the field, leveraging CNNs~\cite{wei2019semi}, Transformers~\cite{Chen2023drsformer}, and state-space models~\cite{zou2024freqmamba} to remove rain from the high-frequency components of rainy images~\cite{fu2017clearing, fu2017removing}. 
Besides, various techniques like multi-scale approaches~\cite{Yasarla2019Uncertaity, Jiang2020MultiScale}, convolutional sparse coding~\cite{wang2020modeldriven}, adversarial learning~\cite{wang2018perceptual, zhang2020image}, transfer learning~\cite{wei2019semi}, and self-attention mechanisms~\cite{zamir2022restormer, Chen2023drsformer, sun2025restoring} have been incorporated. 
Although single-image deraining methods can be adapted for video deraining, they often fail to fully utilize temporal information. 
As a result, specialized deraining models tailored for video data are required to achieve optimal performance.

\noindent\textbf{Video Deraining}
\looseness-1
We can similarly express the relationship for a rainy video:
\begin{equation}
    \mathbf{I}_n = \mathbf{B}_n + \mathbf{R}_n,
\end{equation}
where $\mathbf{I}_n$, $\mathbf{B}_n$, and $\mathbf{n}_t$ denote the respective components of the $n$-th frame in a rainy video. 
Garg and Nayar were pioneers in video deraining, developing a rain detector based on photometric appearance~\cite{Garg2004DetectionAR,Garg2005ICCV,garg2007visionRain}. 
Drawing inspiration from their work, numerous video deraining methods have since emerged, exploring various sources of prior knowledge such as chromatic properties~\cite{zhang2006rain}, rain shape~\cite{Bossu2011Rain}, low rankness~\cite{Chen2013LowRankModel, Jiang2017NovelTensorBased, Jiang2019FastDerain}, frequency space properties~\cite{Barnum2010Analysis}, rain sparsity and dynamics~\cite{Li2018VideoRainStreak, Jiang2019FastDerain}, and temporal correlations~\cite{You2016Adherent}. 
Techniques such as local phase congruency~\cite{Santhaseelan2015Utilizing}, Gaussian mixture models~\cite{Wei2017Should}, tensor factorization~\cite{Jiang2017NovelTensorBased, Kim2015VideoDeraining, Ren2017VideoDesnowing}, and statistical models~\cite{Barnum07spatio-temporalfrequency} have been employed. 
However, methods relying on manually tuned hyper-parameters and heavy computation often struggle to efficiently capture the complex distribution of rain.

Recent advancements in video deraining have been driven largely by deep learning-based methods. 
Various frameworks have been developed, including CNNs~\cite{Chen2018RobustCNN,Liu2019D3RNet} recurrent networks~\cite{Liu2018Erase, Yang2019Frame, Yang2022MFGAN}, self-learning approaches~\cite{Yang2020Self, yang2022learning}, dynamic rain generators~\cite{Yue2021Semi}, a combination of convolutional and recurrent networks~\cite{zhang2022enhanced}, multi-patch progressive learning~\cite{sun2023event}, and state-space models~\cite{wu2024rainmamba}.
However, existing data-driven methods continue to face challenges in bridging the gap between synthetic and real-world data. 
Furthermore, they often overlook the application of real-world rain streak removal in videos for downstream tasks, such as object detection and tracking.

\noindent\textbf{Vision Mamba.}  
\looseness-1  
State-space models (SSMs)~\cite{smith2022simplified,gu2021mamba,lieber2024jamba,zhu2024visionmamba,sun2024hybrid,xie2024fusionmamba}, such as Mamba~\cite{gu2023mamba,hatamizadeh2024mambavision}, have recently demonstrated their efficiency in long-range modeling.  
While several works have applied Mamba to vision tasks~\cite{yamashita2024image,guo2024mambair,zhen2024freqmamba,li2024fouriermamba}, attention-based and CNN models are still reported to outperform Mamba in local and short-sequence modeling~\cite{yu2024mambaout}.  
To enhance global modeling capabilities, existing Mamba-based deraining methods~\cite{zhen2024freqmamba,li2024fouriermamba,yamashita2024image} leverage frequency-domain transformations, applying computations in the Fourier space to improve performance.

\section{Method}




\subsection{Video Deraining Pipeline}

\begin{figure}[t]
  \centering
  \subfloat[Rainy $\mathbf{I}_n$]{\label{fig:warp_example-rainy1}\includegraphics[trim=0 {0.3\linewidth} 0 0,clip,width=0.33\linewidth]{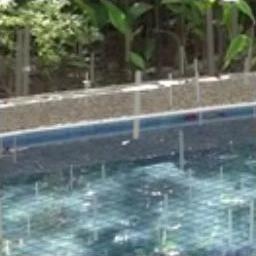}}
  \hfill
  \subfloat[Flow by $\mathcal{F}$]{\label{fig:warp_example-flow1}\includegraphics[trim=0 {0.43\linewidth} 0 0,clip,width=0.33\linewidth]{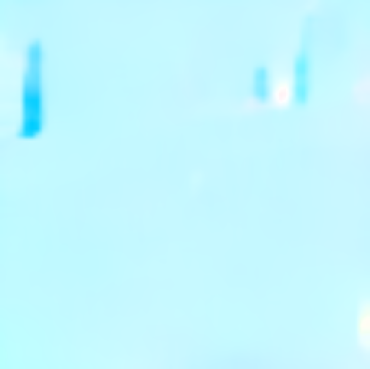}}
  \hfill
  \subfloat[Warped by (b)]{\label{fig:warp_example-warp1}\includegraphics[trim=0 {0.3\linewidth} 0 0,clip,width=0.33\linewidth]{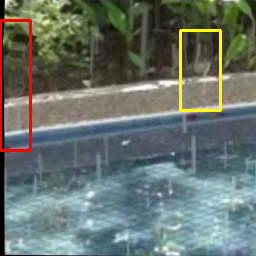}}

  \centering
  \subfloat[Rainy $\mathbf{I}_{n+1}$]{\label{fig:warp_example-rainy2}\includegraphics[trim=0 {0.3\linewidth} 0 0,clip,width=0.33\linewidth]{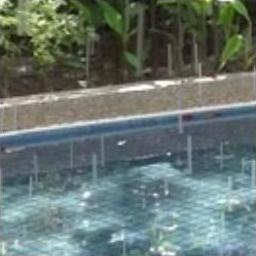}}
  \hfill
  \subfloat[Flow by ours]{\label{fig:warp_example-flow2}\includegraphics[trim=0 {0.43\linewidth} 0 0,clip,width=0.33\linewidth]{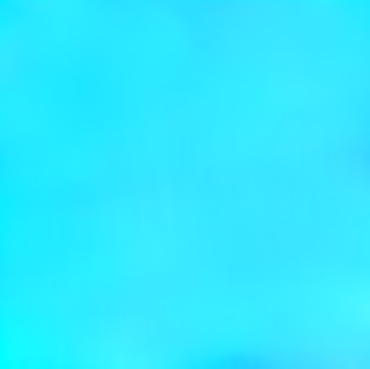}}
  \hfill
  \subfloat[Warped by (e)]{\label{fig:warp_example-warp2}\includegraphics[trim=0 {0.3\linewidth} 0 0,clip,width=0.33\linewidth]{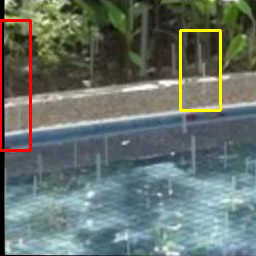}}
\vspace{-3mm}
  \caption{
  Given two adjacent rainy frames (\ref{fig:warp_example-rainy1} \& \ref{fig:warp_example-rainy2}), rain degradation leads to errors in optical flow estimation and subsequent frame alignment (\ref{fig:warp_example-flow1} \& \ref{fig:warp_example-warp1}).
  In contrast, our degradation-free multi-frame estimation pipeline facilitates more accurate warping.}
  \label{fig:warp_example}
  \vspace{-6mm}
\end{figure}

\vspace{1mm}
\noindent\textbf{Multi-Frame Optical Flow Estimation.}
Previous optical flow-aware video deraining methods~\cite{Yang2020Self,Yan_2021_CVPR} require multiple pairwise optical flow estimations to warp and align several neighboring frames, leading to significant latency during flow prediction. 
Additionally, optical flow directly estimated from degraded frames may be unreliable~\cite{Yan_2021_CVPR}. 
Inspired by knowledge distillation~\cite{Hinton2015DistillingTK,sun2024logit}, we adopt a pipeline using degradation-free multi-frame optical flow transfer. 
Given a sequence of $N$ rainy frames $\mathbf{I}_1, ..., \mathbf{I}_N$ and their corresponding ground-truths, $\textbf{B}_1,...,\textbf{B}_N$, a pre-trained flow estimator, \(\mathcal{F}\), predicts the pairwise optical flow for the clean pairs,
\begin{equation}
    \mathbf{\tilde O}_{n\rightarrow c} \!=\! \mathcal{F} \left(\mathbf{B}_n, \mathbf{B}_c\right),\ n\in\{1,...,c-1,c+1,...,N\},
\end{equation}
where \(c\) denotes the index of the current center frame, and \(\mathbf{\tilde O}_{n\rightarrow c}\in\mathbb{R}^{H\times W\times 2}\) represents the optical flow of the \(n\)-th frame towards the center frame.
We define our multi-frame flow estimator as $f$, which
estimates multi-frame flows by
\begin{equation}
    \O_{1\rightarrow c}, ..., \O_{N\rightarrow c} = f \left(\mathbf{I}_1, ..., \mathbf{I}_N\right).
\end{equation}
The multi-frame flow estimator is trained by minimizing the optical flow transfer loss, which is formulated as:
\begin{equation}\label{eq:distillation_loss}
    \mathcal{L}_{flow} = \sum_{n=1, n\ne c}^{N}\|\O_{n\rightarrow c} - \mathbf{\tilde O}_{n\rightarrow c}\|_1.
\end{equation}
In this manner, a single model forward pass generates multiple optical flows simultaneously, effectively reducing runtime. 
Furthermore, this approach helps mitigate the degradation of flow inference caused by rain distortions, as shown in Fig.~\ref{fig:warp_example}. 
With the trained multi-frame flow estimation, we can efficiently align neighboring frames for the deraining task.
For both multi-frame flow estimation and subsequent video deraining, we use the same structure of our VDMamba, except that the projected output channel dimension is \(2\times (N-1)\) for flow estimation and \(3\) for deraining.
We also include the restored previous frames along with the rainy frames, though this detail is omitted for brevity.

\begin{figure*}[t]
  \centering
   \includegraphics[width=1\linewidth]{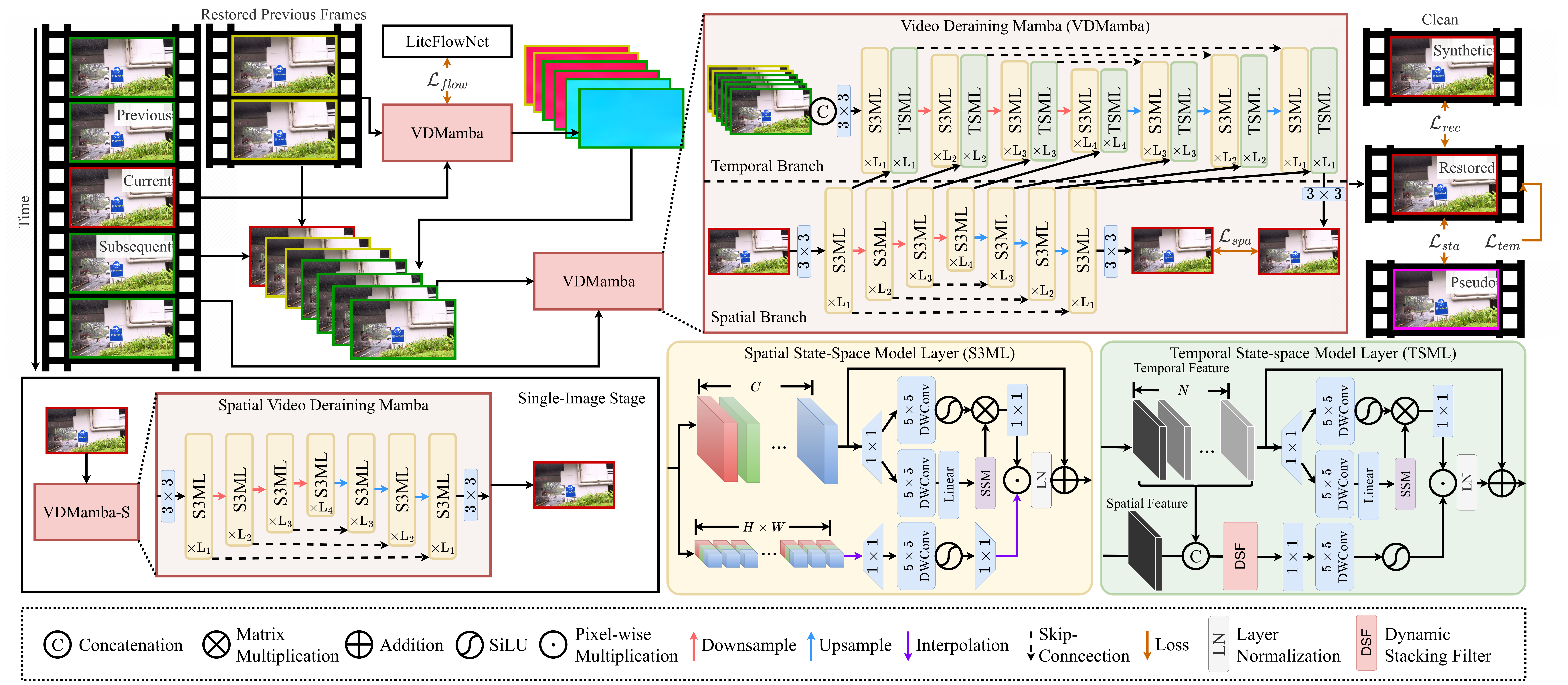}
\vspace{-7mm}
   \caption{The architecture of our proposed VDMamba for video deraining, consisting of spatial state-space model layers (S3ML) for single-frame feature extraction and temporal state-space model layers (TSML) for multi-frame feature fusion.
   Due to the modular design of the spatial branch, the sub-model, VDMamba-S, which contains only S3ML, is also capable of performing single-image deraining.
   }
   \label{fig:arch}
   \vspace{-2mm}
\end{figure*}

\vspace{1mm}
\noindent\textbf{Overall Structure.}
The overall architecture of our VDMamba is shown in Fig.~\ref{fig:arch}.
It consists of two distinct branches, one for spatial feature extraction and the other for temporal feature fusion. 
Both branches follow a U-Net~\cite{Ronneberger2015UNetCN} architecture, comprising four levels of feature transformation through upsampling and downsampling. 
The encoders and decoders are connected through skip connections using residual learning.
The network includes two core components: the spatial state-space model layer (S3ML) and the temporal state-space model layer (TSML).
Given the spatial and temporal branch features, \(\mathbf{F}^i_S\in\mathbb{R}^{H\times W\times C_S}\) and \(\mathbf{F}^i_T\in\mathbb{R}^{H\times W\times C_T}\) at the \(i\)-th level, with the height \(H\), the width \(W\), and the channels of \(C_S\) and \(C_T\), we have
\begin{equation}
\begin{split}
    &\mathbf{F}^{i}_S = \text{S3ML}(\mathbf{F}^{i-1}_S),\ \mathbf{F}^{i}_T = \text{S3ML}(\mathbf{F}^{i-1}_T),\\
    &\mathbf{F}^{i+1}_T = \text{TSML}(\mathbf{F}^i_S, \mathbf{F}^i_T).
\end{split}
\end{equation}
At the beginning and end of the branches, two convolutions with a kernel size of \(3\times 3\) are used for patch embedding and output projection, respectively.

\vspace{1mm}
\noindent\textbf{Spatial State-Space Model Layer} is designed for spatial feature extraction.
Given a spatial feature \(\mathbf{F}_S^i\), we can express S3ML as:
\begin{equation*}
\begin{split}
    &\mathbf{F}_{S,1}^i, \mathbf{F}_{S,2}^i = \mathrm{C_{1}}(\mathbf{F}_S^i) ,\\
    & \mathbf{F}_{S,1,2}^i= \mathrm{C_{1}}\!\left(\sigma(\mathrm{C^{D}_{5}}(\mathbf{F}_{S,1}^i)) \otimes \text{SSM}(\mathrm{L}(\mathrm{C^{D}_{5}}(\mathbf{F}_{S,2}^i)))\!\right) ,\\
\end{split}
\end{equation*}
\begin{equation*}
\begin{split}
    &\mathbf{F}_{S,3}^i = \mathcal{I}\left(\mathrm{C_{1}}\left(\sigma\left(\mathrm{C^{D}_{5}}\left(\mathrm{C_{1}}\left(\mathcal{I}(\mathbf{F}_S^i)\right)\right)\right)\right)\right) ,\\
    & \mathbf{F}_{S}^{i+1} = \text{LN}(\mathbf{F}_{S,1,2}^i \odot \mathbf{F}_{S,3}^i) + \mathbf{F}_S^i, \\
\end{split}
\end{equation*}
where \(\mathrm{C_{1}}\), \(\mathrm{C^{D}_{5}}\), and \(\text{L}\) denote point-wise convolution of a kernel size of \(1\times 1\), depth-wise convolution with a kernel size of \(5\times 5\), and a linear layer, respectively. 
LN represents layer normalization, while \(\sigma\) is the SiLU~\cite{elfwing2018sigmoid} activation function. 
The operations \(\odot\) and \(\otimes\) refer to element-wise multiplication and matrix multiplication, respectively.
\(\mathcal{I}\) represents the interpolation process.
The selective scan module (\(\text{SSM}\)) operates along the channel dimension to capture global dependencies, while we refine the local information by computing an additional \(\mathbf{F}_{S,3}^i\) feature.

\vspace{1mm}
\noindent\textbf{Temporal State-space Model Layer} is designed for temporal feature fusion.
Given a spatio-temporal feature \(\mathbf{F}_T^i\), we can express TSML as:
\begin{equation}
\begin{split}
    &\mathbf{F}_{T,1}^i, \mathbf{F}_{T,2}^i = \mathrm{C_{1}}(\mathbf{F}_T^i) ,\\
    & \mathbf{F}_{T,1,2}^i= \mathrm{C_{1}}\!\left(\sigma(\mathrm{C^{D}_{5}}(\mathbf{F}_{T,1}^i)) \otimes \text{SSM}(\mathrm{L}(\mathrm{C^{D}_{5}}(\mathbf{F}_{T,2}^i)))\!\right) ,\\
    &\mathbf{F}_{T,3}^i = \sigma\left(\mathrm{C^{D}_{5}}\left(\mathrm{C_{1}}\left(\text{DSF}([\mathbf{F}_S^i;\mathbf{F}_T^i])\right)\right)\right) ,\\
    & \mathbf{F}_{T}^{i+1} = \text{LN}(\mathbf{F}_{T,1,2}^i \odot \mathbf{F}_{T,3}^i) + \mathbf{F}_T^i, \\
\end{split}
\end{equation}
where \([ \cdot\ ;\cdot ]\) denotes the concatenation operation. 
The SSM operation is performed along the temporal dimension to integrate multi-frame information. 
DSF refers to the proposed dynamic stacking filter, whose derivation is provided below.

\subsection{Dynamic Stacking Filter}

Given the feature candidates of aligned neighboring frames, conventional operations, such as a convolution layer with consistent parameters, struggle to selectively integrate pixel-adaptive features~\cite{aloysius2017review}.
Inspired by the benefits of stacking techniques in video deraining, as illustrated in Fig.~\ref{fig:stacking}, we propose a dynamic stacking filter to adaptively select a statistics from the mean, median, and extreme values (maximum and minimum) to fuse pixels across the temporal dimension. 
However, among the statistics, only the mean is a differentiable value. 
To address the indifferentiability of the others, we define their differentiable approximations.

\vspace{1mm}
\noindent\textbf{Differentiable Approximation to Max and Min.}
Given a set of pixels $G=\{x_1,...,x_N\}$, if we know the index of the minimum, i.e.,
\begin{equation}\label{eq:indexMin}
    \smoothArgMin(G)_n = 
    \begin{cases} 
        \hspace{9.4mm} 0&\text{, if $x_n \ne \min(G)$}, \\
        \frac{1}{\left|\{x|x=\min(G)\}\right|}&\text{, otherwise,}
    \end{cases}
\end{equation}
then we can compute the minimum by 
\vspace{-0.5mm}\begin{equation}
    \smoothMin(G) = \sum_{n=1}^N  x_n \smoothArgMin(G)_n. 
\end{equation}
The index of the minimum in Eq.~\ref{eq:indexMin} can be approximated using a generalized softmax function,
\vspace{-0.5mm}\begin{align}
    \smoothArgMax(G;a)_n &\approx \frac{ e^{ax_n}}{\sum_{j=1}^N e^{ax_j}}, \\
    \smoothArgMin(G) &= \smoothArgMax(\{-x_1, ..., -x_N\}), 
\end{align}
where $a$ is a scaling coefficient, analogous to the temperature used in distillation~\cite{Hinton2015DistillingTK}, and the subscript of \(n\) denotes the \(n\)-th element of function's output. 
The approximation converges to equality as $a\rightarrow \infty$. 
Thus, a differentiable expression for the approximated minimum can be  
\begin{equation}
    \smoothMin(G;a) \approx \frac{\sum_{n=1}^N x_n e^{-ax_n}}{\sum_{j=1}^N e^{-ax_j}}.
\end{equation}
Similarly, the differentiable estimate of maximum is obtained by computing $\sum_{n=1}^Nx_n\smoothArgMax(G)_n$.

\vspace{1mm}
\noindent\textbf{Differentiable approximation to median.}
We begin with the fact that the median is a solution to minimizing the mean absolute deviation, i.e., 
\vspace{-1mm}\begin{theorem}\label{theorem:1}
Given a set of values $G=\left\{x_n\right\}_{n=1}^N$, its median is a solution to minimizing the mean absolute deviation, as expressed by the equation:
\vspace{-1mm}\begin{equation}\label{eq:medianMAD}
    \med(G) \in \mathop{\arg\min}\limits_{x} \frac{1}{N}\sum_{n=1}^N|x - x_n|,
\end{equation}
where $\med(\cdot)$ denotes the median of a set.
\end{theorem}
The proof is provided in the \textit{Supplements}.
Similar to the definition of $\smoothArgMin$ in Eq.~\ref{eq:indexMin}, we can define the index of the median with the assistance of Eq.~\ref{eq:medianMAD}, as follows:
%
\begin{equation}
    \hspace{-2mm}
    \smoothArgMed(G)_n = \smoothArgMin \left( \left\{\frac{1}{N}\sum_{j=1}^N\left|x_n - x_j\right| \right\}_{n=1}^N\right)\!.
    \hspace{-2mm}
\end{equation}
Thus, we can obtain the differentiable approximation to the median as follows:
\vspace{-2mm}\begin{equation}\label{eq:smoothMedian}
\begin{split}
    \med(G;a) &= \sum_{n=1}^N x_n\smoothArgMed(G)_n \\
    &\approx \frac{\sum_{n=1}^N x_n e^{a \frac{1}{N}\sum_{j=1}^N|x_n - x_j|}}{\sum_{k=1}^N e^{a \frac{1}{n}\sum_{j=1}^N|x_k - x_j|}}.
\end{split}
\end{equation}
Note that the absolute value function is differentiable except when the input equals zero. 
It can also be approximated by $|x|\approx \smoothArgMax(\{x, -x\}; a)$. 
This approximation converges to equality as $a\rightarrow \infty$.


\vspace{1mm}
\noindent\textbf{Dynamic stacking filter.}
To design a adaptive stacking filter that can degrade to four options of mean, median, maximum and minimum values, we introduce a gate coefficient $b$ in Eq.~\ref{eq:smoothMedian}, resulting in the following expression: 
\begin{equation}
\begin{split}
    &\agg(G; a,b) = \frac{\sum_{n=1}^N x_n e^{a \frac{1}{n}\sum_{j=1}^n|x_n - bx_j|}}{\sum_{k=1}^n e^{a \frac{1}{n}\sum_{j=1}^n|x_k - b x_j|}} \\
    =& \left\{
\begin{split}
    \mean(G) &\text{, if $a\rightarrow 0$, $b\rightarrow \infty$, } \\
    \med(G) &\text{, if $a\rightarrow \infty$, $b\rightarrow 1$, } \\
    \smoothMax(G) &\text{, if $a\rightarrow \infty$, $b\rightarrow 0$, $x_n\ge 0$,} \\
    \smoothMin(G) &\text{, if $a\rightarrow -\infty$, $b\rightarrow 0$, $x_n\ge 0$}.
\end{split}
\right.
\end{split}
\end{equation}
%
The cases where \(x_n < 0\) are symmetric and thus omitted for brevity.
In the case of matrix inputs \([\mathbf{F}_S^i, \mathbf{F}_T^i]\), the filter \(\agg\) is applied to each pixel, with constant coefficients becoming learnable matrix parameters $\a, \b\in \mathbb{R}^{H \times W}$.



\subsection{Loss functions for Semi-Supervision}

We train the model for multi-frame optical flow estimation using the optical flow transfer loss in Eq.~\ref{eq:distillation_loss}.
To train VDMamba for video deraining, we employ three loss terms in a semi-supervision manner.
Given the ground truth, the first two loss terms represent reconstruction losses in the form of the \(L_1\) norm:
%
\begin{equation}\label{eq:reconstruction_loss}
    \mathcal{L}_{rec} = \left\|\mathbf{B}_c - \mathbf{\tilde B}_c\right\|_1, \quad \mathcal{L}_{spa} = \left\| \mathbf{\overline B}_c - \mathbf{\tilde B}_c \right\|_1,
\end{equation}
where \(\mathbf{\tilde B}_c\) and \(\mathbf{\overline B}_c\) denote the current frames restored by our dual-branch model and the spatial-branch-only model, respectively. 
The loss term \(\mathcal{L}_{spa}\) is designed to enhance the reconstruction capacity of the spatial branch.
In the case of unpaired real rainy data, we construct pseudo ground-truths using a masked median stacking technique designed to remove moving objects.
Specifically, we first obtain aligned neighboring frames \(\mathbf{\tilde I}_n\) by
\begin{equation}\label{eq:warped_frames}
\mathbf{\tilde I}_n = \mathcal{W}\left(\mathbf{I}_n; \O_{n\rightarrow c}\right) \text{, for \(n=1,...,N\),}
\end{equation}
where $\mathcal{W}(\cdot; \cdot)$ is the backward warping function. 
%
With the aligned frame, we compute the median of the consecutive aligned frames, denoted by
\begin{equation}
    \mathbf{M}_c = \med\left(\left\{\mathbf{I}_c, \mathbf{\tilde I}_1,...\mathbf{\tilde I}_N \right\}\right).
\end{equation}
However, directly applying the temporal median to the aligned sequence may result in mismatches caused by occlusions and mistaken removal of moving backgrounds instead of rain streaks. 
To mitigate this, we first crop the frames into $P^2$ sub-patches, each with width $\frac{W}{P}$ and height $\frac{H}{P}$. 
The corresponding sub-patch of the derained frame, the median frame, and the original central frame are denoted as $\mathbf{\hat B}_{c,p}$, $\mathbf{\hat M}_{c,p}$, and $\mathbf{\hat I}_{c,p}$ for $p=\{1,...,P^2\}$. 
Due to the sparsity of rain streaks, the median sub-patch $\hat M_{t,i}$ is adopted as a pseudo-clean image when the percentage of unchanged pixels compared to $\mathbf{\hat I}_{c,p}$ exceeds a threshold $\theta$.
\begin{equation}\label{eq:stacking_patch_index_mask}
\begin{split}
    &\mathcal{M}\left(\mathbf{\hat M}_{c,p},  \mathbf{\hat I}_{c,p}\right) \\
    = &\left\{
    \begin{split}
        &1 \text{, if $\frac{P^2}{WH}\sum_{j,k} \mathbbm{1}_{\left\{ \left|\mathbf{\hat M}_{c,p}[j,k] - \mathbf{\hat I}_{c,p}[j,k]\right|\le \delta\right\}} \ge \theta$, } \\
        &0\text{, otherwise},
    \end{split}\right.
\end{split}
\end{equation}
where $\delta$ is a slack variable that controls the tolerance for the decision condition, \(\mathcal{M}\) is a masking function to select rain-free and well-aligned patches, and \(\mathbbm{1}_{\{\textit{condition}\}}\) is an indicator function that returns \(1\) if the specific \textit{condition} is met and \(0\) otherwise.
The stacking loss is subsequently computed only on the masked sub-patches:
\begin{equation}\label{eq:stacking_loss}
    \mathcal{L}_{sta} = \sum_{p=1}^{P^2}\mathcal{M} \left(\mathbf{\hat M}_{c,p}, \mathbf{\hat I}_{c,p}\right)\left\|\mathbf{\hat B}_{c,p}- \mathbf{\hat M}_{c,p}\right\|_1.
\end{equation}
To further improve the temporal consistency, we incorporate a temporal loss~\cite{Yang2020Self},
\begin{equation}\label{eq:temporal_loss}
    \mathcal{L}_{tem} = \frac{1}{N-1}\sum_{n=1,n\ne c}^{N}\left\|\mathcal{W}\left(\mathbf{\tilde B}_c; \O_{c\rightarrow n}\right) - \mathbf{I}_n\right\|_1, 
\end{equation}
%
\begin{table*}[htbp]
\footnotesize
\renewcommand{\arraystretch}{0.8}
\setlength{\tabcolsep}{0.5mm}
  \centering
  \caption{Quantitative results of various video deraining methods on \textit{RainSynLight25}~\cite{Liu2018Erase}, \textit{RainSynComplex25}~\cite{Liu2018Erase} and \textit{NTURain}~\cite{Chen2018RobustCNN}.}
  \vspace{-3mm}
    \begin{tabular}{c|c|ccccccccc}
    \toprule
    Dataset & Metrics & Rainy &  FastDeRain~\cite{Jiang2019FastDerain} & SpacCNN~\cite{Chen2018RobustCNN} & J4RNet~\cite{yang2020joint} & DRSformer~\cite{Chen2023drsformer} & ESTINet~\cite{zhang2022enhanced} & MFGAN~\cite{Yang2022MFGAN} & RainMamba~\cite{wu2024rainmamba} & Ours \\
    \midrule
    \multirow{2}[1]{*}{\textit{RainSynLight25}} & PSNR  & 25.23  & 29.42  & 32.78  & 32.96  & 36.84  & 36.12  & 36.99 & 36.74 & \textbf{37.53 } \\
          & SSIM  & 0.9099  &  0.8683  & 0.9239  & 0.9434  & 0.9739  & 0.9631  & 0.9760 & 0.9741 & \textbf{0.9812 } \\
    \midrule
    \multirow{2}[1]{*}{\textit{RainSynComplex25}} & PSNR  & 16.85 & 19.25 & 21.21 & 24.13 & 31.61 & 28.48 & 32.70 & 32.65 & \textbf{32.89} \\
          & SSIM  & 0.6791  & 0.5385  & 0.5854 & 0.7163 & 0.9258 & 0.8242  & 0.9357 & 0.9361 & \textbf{0.9425 } \\
    \midrule

    \multirow{2}[1]{*}{\textit{NTURain}} & PSNR   & 30.41  & 30.30  & 33.11 & 32.14 & 36.93  & 37.48  & 38.92 & 37.87 &  \textbf{39.74 }  \\
    & SSIM & 0.9108  &  0.9274   & 0.9475 & 0.9480  &  0.9591  & 0.9700  & 0.9764 & 0.9738 &\textbf{0.9791} \\
    \bottomrule
    \end{tabular}%
  \label{tab:rainsyn}%
  \vspace{-3mm}
\end{table*}%

where \(\O_{c\rightarrow n} = \mathcal{F}(\tilde B_c, I_n)\) represents the optical flow warping central frame inversely to the neighboring frames.
The total loss for training deraining is calculated by summing the individual loss terms, 
\begin{equation*}
    \mathcal{L} = 
    \begin{cases}
         \mathcal{L}_{rec} + \mathcal{L}_{spa} + \lambda_2 \mathcal{L}_{tem} &\text{, if paired data,}\\
        \lambda_1 \mathcal{L}_{sta} + \mathcal{L}_{spa} + \lambda_2 \mathcal{L}_{tem} &\text{, if unpaired data,} 
    \end{cases}
\end{equation*}
where \(\lambda_1\) and \(\lambda_2\) regulate the balance between the strengths of the supervision terms.

\begin{figure*}
  \centering
  \hspace{-2mm}
  \subfloat[Rainy\label{fig:ra3-2-a}]{
  \begin{minipage}{0.1405\linewidth}
    \centering
    \includegraphics[width=1\linewidth]{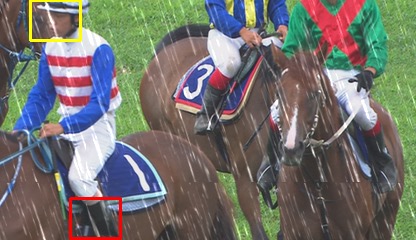}
    \hspace{-0mm}
    \includegraphics[width=0.490\linewidth]{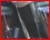}
    \hspace{-1.3mm}
     \includegraphics[width=0.490\linewidth]{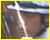}
    
    \centering
    \includegraphics[width=1\linewidth]{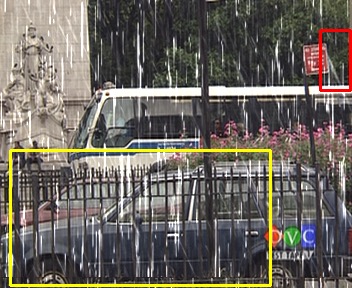}
    \hspace{-0mm}
     \includegraphics[width=1\linewidth]{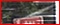}
    
    \centering
    \includegraphics[width=1\linewidth]{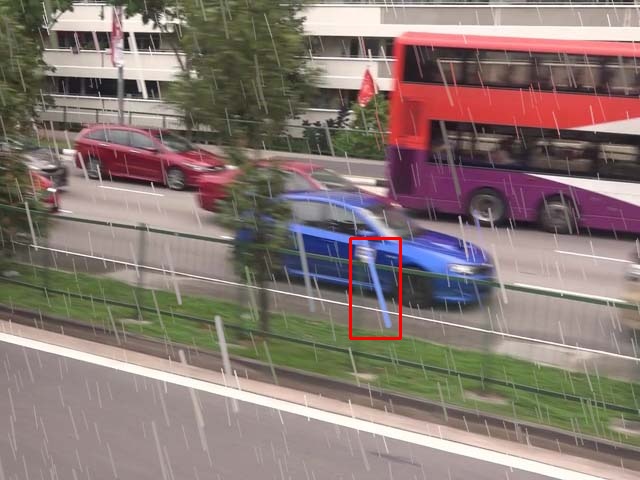}
    \hspace{-0mm}
     \includegraphics[width=1\linewidth]{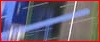}
    \end{minipage}
    }
    \hspace{-2.4mm}
    \subfloat[DRSformer.~\cite{Chen2023drsformer}\label{fig:ra3-2-c}]{
  \begin{minipage}{0.1405\linewidth}
    \centering
    \includegraphics[width=1\linewidth]{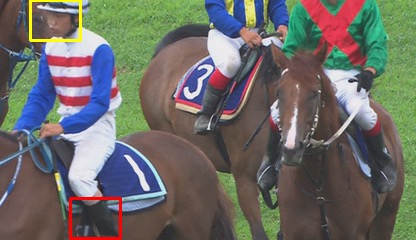}
    \hspace{-0mm}
    \includegraphics[width=0.490\linewidth]{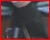}
    \hspace{-1.3mm}
     \includegraphics[width=0.490\linewidth]{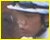}
     
    \centering
    \includegraphics[width=1\linewidth]{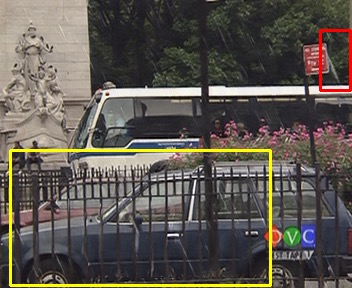}
    \hspace{-0mm}
     \includegraphics[width=1\linewidth]{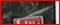}
    
    \centering
    \includegraphics[width=1\linewidth]{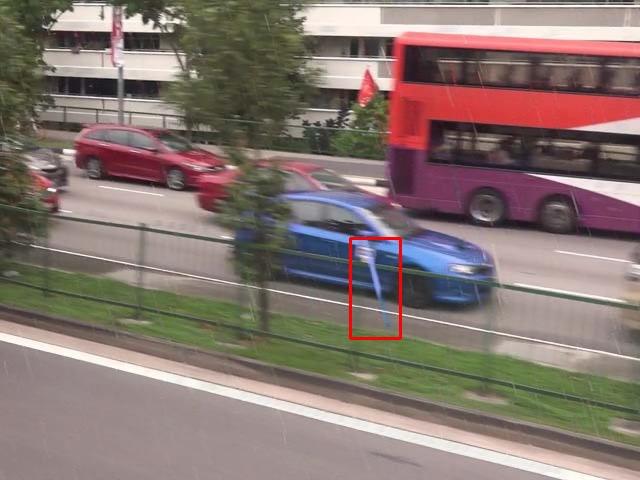}
    \hspace{-0mm}
     \includegraphics[width=1\linewidth]{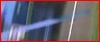}
    \end{minipage}
    }
    \hspace{-2.4mm}
    \subfloat[ESTINet~\cite{zhang2022enhanced}\label{fig:ra3-2-d}]{
  \begin{minipage}{0.1405\linewidth}
    \centering
    \includegraphics[width=1\linewidth]{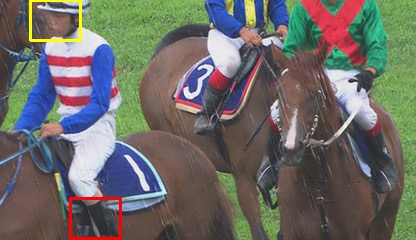}
    \hspace{-0mm}
    \includegraphics[width=0.490\linewidth]{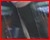}
    \hspace{-1.3mm}
     \includegraphics[width=0.490\linewidth]{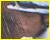}
     
    \centering
    \includegraphics[width=1\linewidth]{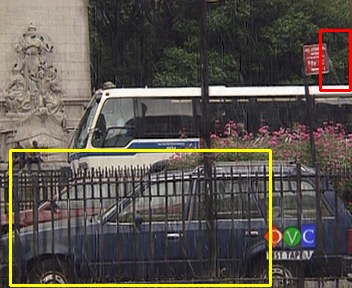}
    \hspace{-0mm}
     \includegraphics[width=1\linewidth]{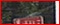}
    
    \centering
    \includegraphics[width=1\linewidth]{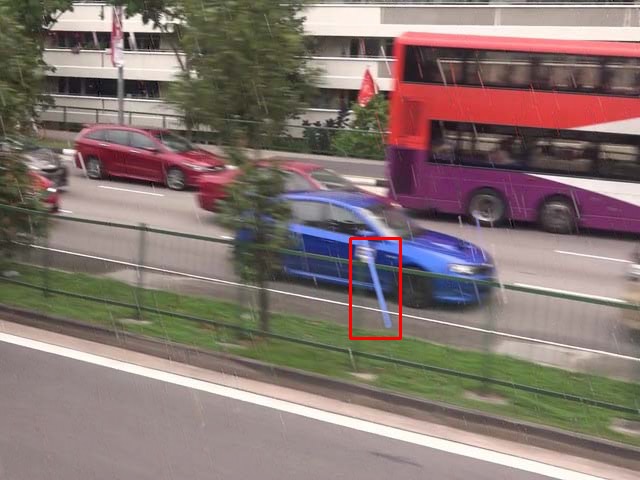}
    \hspace{-0mm}
     \includegraphics[width=1\linewidth]{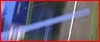}
    \end{minipage}
    }
    \hspace{-2.4mm}
    \subfloat[MFGAN~\cite{Yang2022MFGAN}\label{fig:ra3-2-e}]{
  \begin{minipage}{0.1405\linewidth}
    \centering
    \includegraphics[width=1\linewidth]{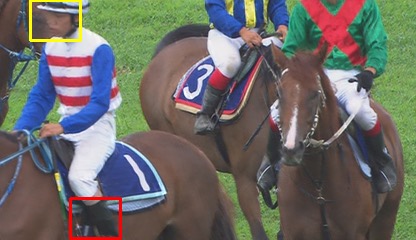}
    \hspace{-0mm}
    \includegraphics[width=0.490\linewidth]{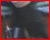}
    \hspace{-1.3mm}
     \includegraphics[width=0.490\linewidth]{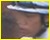}
     
    \centering
    \includegraphics[width=1\linewidth]{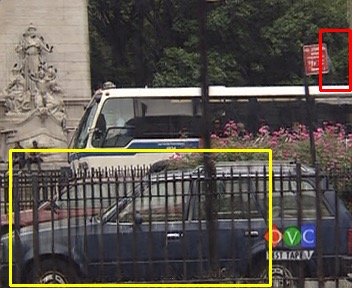}
    \hspace{-0mm}
     \includegraphics[width=1\linewidth]{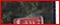}
    
    \centering
    \includegraphics[width=1\linewidth]{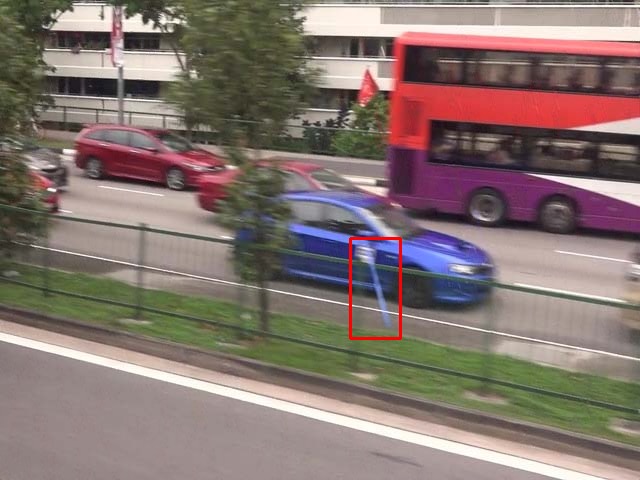}
    \hspace{-0mm}
     \includegraphics[width=1\linewidth]{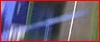}
    \end{minipage}
    }
    \hspace{-2.4mm}
    \subfloat[RainMamba~\cite{wu2024rainmamba}\label{fig:ra3-2-f}]{
  \begin{minipage}{0.1405\linewidth}
    \centering
    \includegraphics[width=1\linewidth]{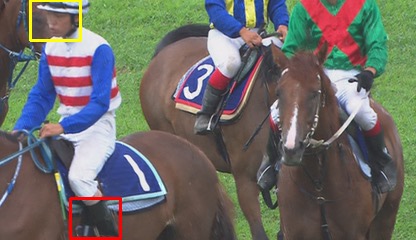}
    \hspace{-0mm}
    \includegraphics[width=0.490\linewidth]{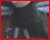}
    \hspace{-1.3mm}
     \includegraphics[width=0.490\linewidth]{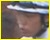}
     
    \centering
    \includegraphics[width=1\linewidth]{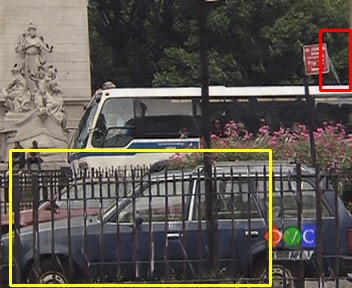}
    \hspace{-0mm}
     \includegraphics[width=1\linewidth]{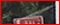}
    
    \centering
    \includegraphics[width=1\linewidth]{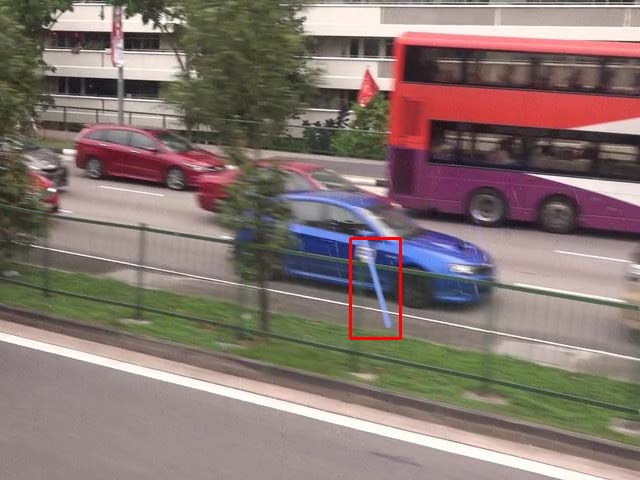}
    \hspace{-0mm}
     \includegraphics[width=1\linewidth]{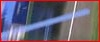}
    \end{minipage}
    }
    \hspace{-2.4mm}
    \subfloat[Ours\label{fig:ra3-2-b}]{
  \begin{minipage}{0.1405\linewidth}
    \centering
    \includegraphics[width=1\linewidth]{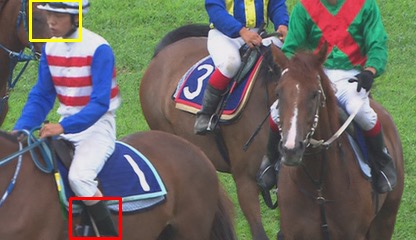}
    \hspace{-0mm}
    \includegraphics[width=0.490\linewidth]{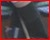}
    \hspace{-1.3mm}
     \includegraphics[width=0.490\linewidth]{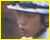}
     
    \centering
    \includegraphics[width=1\linewidth]{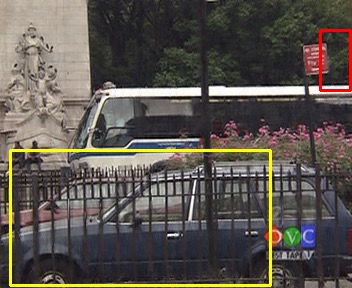}
    \hspace{-0mm}
     \includegraphics[width=1\linewidth]{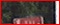}
    
    \centering
    \includegraphics[width=1\linewidth]{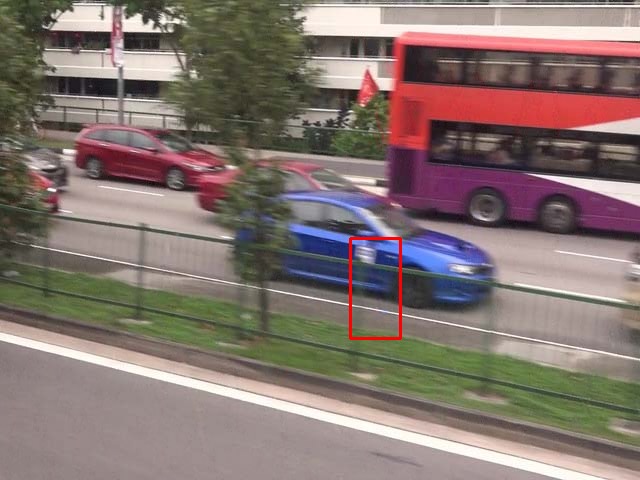}
    \hspace{-0mm}
     \includegraphics[width=1\linewidth]{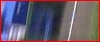}
    \end{minipage}
    }
    \hspace{-2.4mm}
    \subfloat[Clean\label{fig:ra3-2-g}]{
  \begin{minipage}{0.1405\linewidth}
    \centering
    \includegraphics[width=1\linewidth]{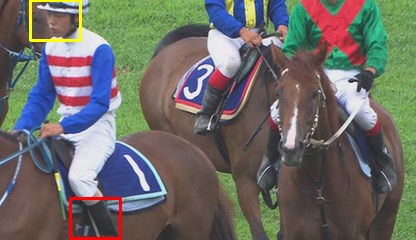}
    \hspace{-0mm}
    \includegraphics[width=0.490\linewidth]{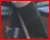}
    \hspace{-1.3mm}
     \includegraphics[width=0.490\linewidth]{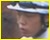}
     
    \centering
    \includegraphics[width=1\linewidth]{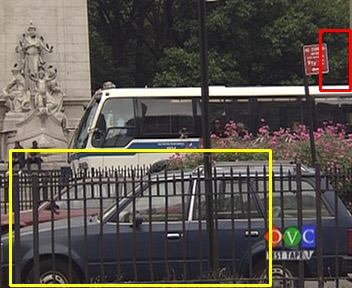}
    \hspace{-0mm}
     \includegraphics[width=1\linewidth]{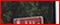}
    
    \centering
    \includegraphics[width=1\linewidth]{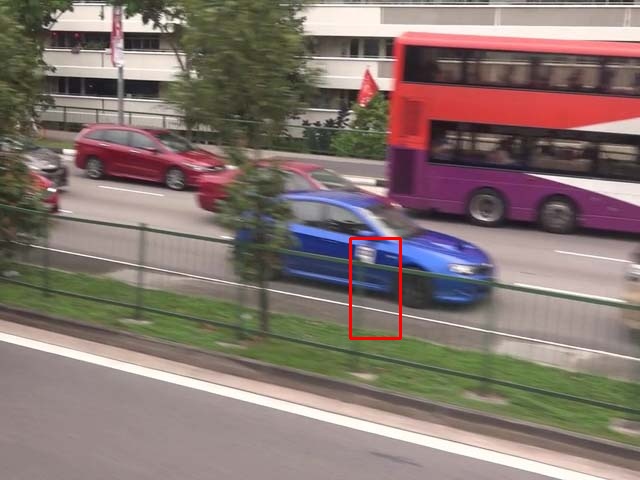}
    \hspace{-0mm}
     \includegraphics[width=1\linewidth]{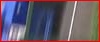}
    \end{minipage}
    }
    \vspace{-3mm}
  \caption{The qualitative comparison among the existing deraining methods on synthetic datasets \textit{RainSynLight25}~\cite{Liu2018Erase}, \textit{RainSynComplex25}~\cite{Liu2018Erase} and \textit{NTURain}~\cite{Chen2018RobustCNN}. 
  Please zoom in for better view.
  }
  \label{fig:realra3}
  \vspace{-2mm}
\end{figure*}

\subsection{Real-World Rainy Task Benchmark}
In addition to enhancing visual quality, deraining models hold practical potential for restoring degraded videos, enabling more effective performance in downstream high-level vision tasks. 
To this end, we introduce the real-world rainy video object detection and tracking benchmark, referred to as RVDT.
Specifically, we curated 57 long video clips, totaling 10,258 frames, selected from the internet, each exhibiting noticeable rain streak degradation. 
We have annotated 33,077 objects for detection and 408 unique objects for tracking. 
Detailed statistics and examples are provided in the \textit{Supplements}.

\begin{figure*}
  \centering
  \hspace{-2mm}
  \subfloat[Rainy\label{fig:real-3-a}]{
  \begin{minipage}{0.1405\linewidth}
    \centering
    \includegraphics[width=1\linewidth]{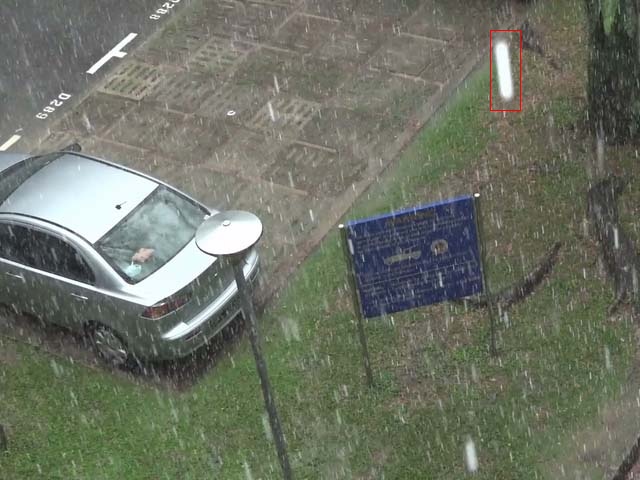}
    \hspace{-0mm}
    \includegraphics[width=1\linewidth]{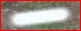}
    
    \centering
    \includegraphics[width=1\linewidth]{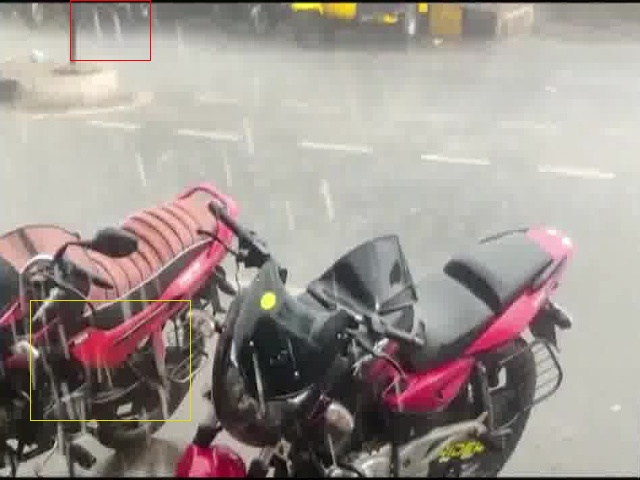}
    \hspace{-0mm}
    \includegraphics[width=0.490\linewidth]{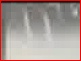}
    \hspace{-1.3mm}
     \includegraphics[width=0.490\linewidth]{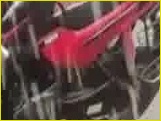}
    \end{minipage}
    }
    \hspace{-2.4mm}
    \subfloat[FastDeRain~\cite{Jiang2019FastDerain}\label{fig:real-3-b}]{
  \begin{minipage}{0.1405\linewidth}
    \centering
    \includegraphics[width=1\linewidth]{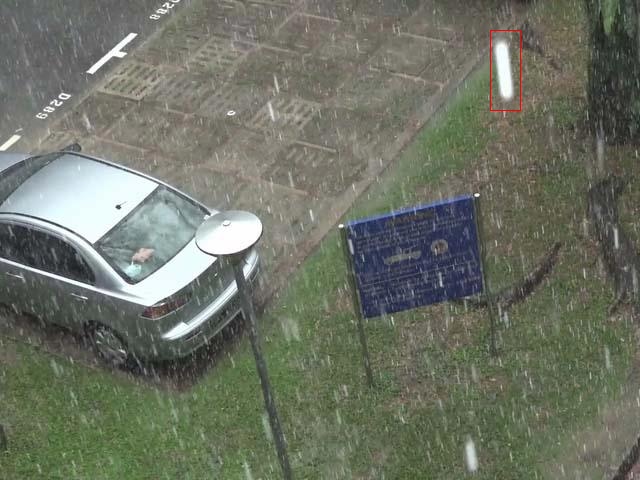}
    \hspace{-0mm}
    \includegraphics[width=1\linewidth]{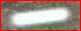}
     
    \centering
    \includegraphics[width=1\linewidth]{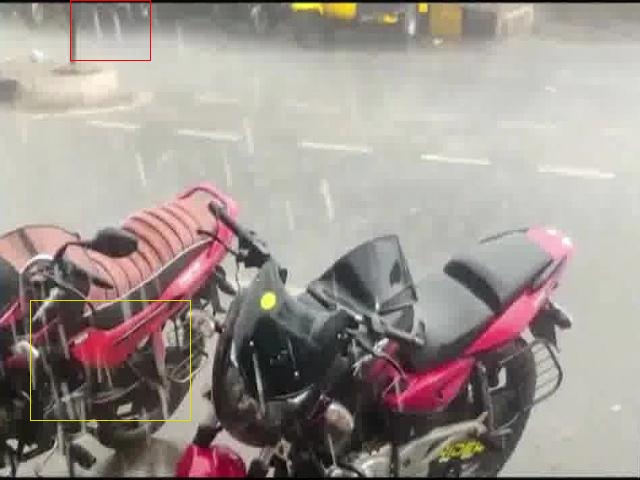}
    \hspace{-0mm}
    \includegraphics[width=0.490\linewidth]{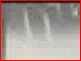}
    \hspace{-1.3mm}
     \includegraphics[width=0.490\linewidth]{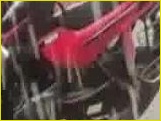}
    \end{minipage}
    }
    \hspace{-2.4mm}
    \subfloat[DRSformer.~\cite{Chen2023drsformer}\label{fig:real-3-c}]{
  \begin{minipage}{0.1405\linewidth}
    \centering
    \includegraphics[width=1\linewidth]{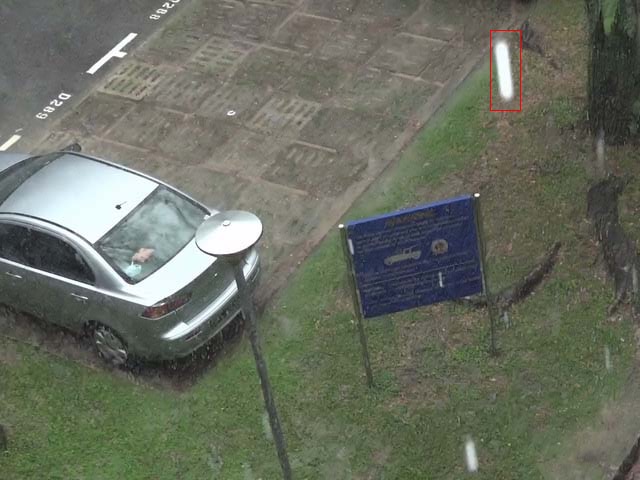}
    \hspace{-0mm}
    \includegraphics[width=1\linewidth]{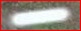}
     
    \centering
    \includegraphics[width=1\linewidth]{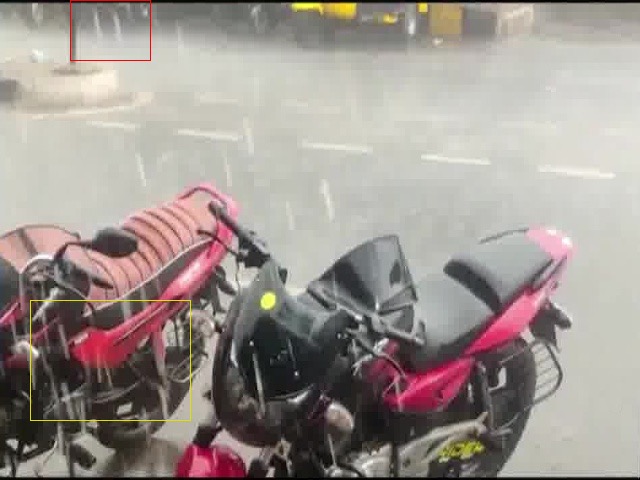}
    \hspace{-0mm}
    \includegraphics[width=0.490\linewidth]{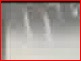}
    \hspace{-1.3mm}
     \includegraphics[width=0.490\linewidth]{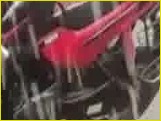}
    \end{minipage}
    }
    \hspace{-2.4mm}
    \subfloat[ESTINet~\cite{zhang2022enhanced}\label{fig:real-3-d}]{
  \begin{minipage}{0.1405\linewidth}
    \centering
    \includegraphics[width=1\linewidth]{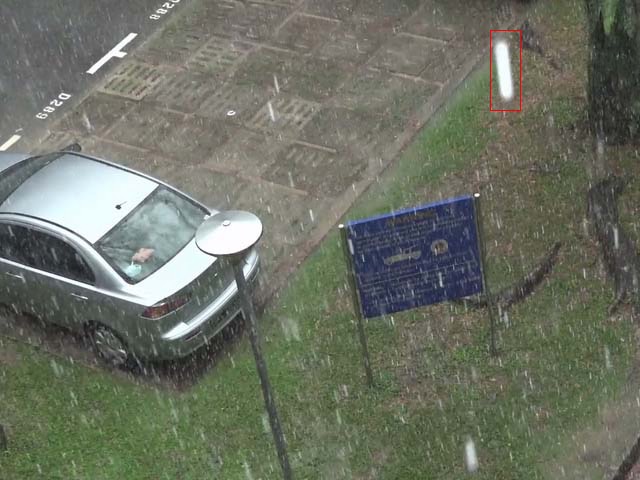}
    \hspace{-0mm}
    \includegraphics[width=1\linewidth]{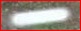}
     
    \centering
    \includegraphics[width=1\linewidth]{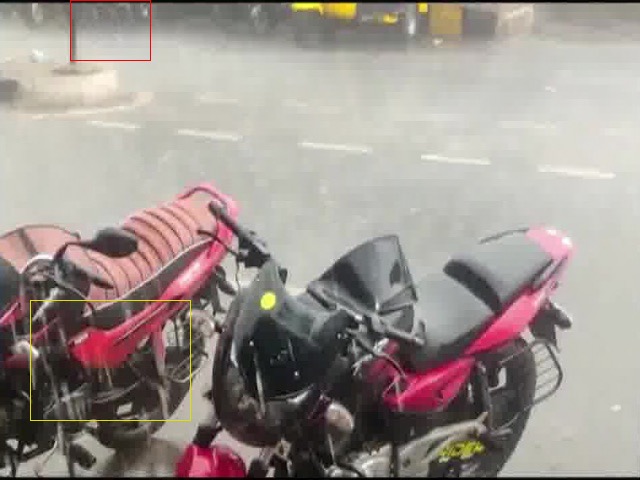}
    \hspace{-0mm}
    \includegraphics[width=0.490\linewidth]{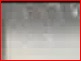}
    \hspace{-1.3mm}
     \includegraphics[width=0.490\linewidth]{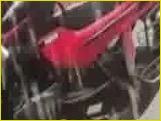}
    \end{minipage}
    }
    \hspace{-2.4mm}
    \subfloat[MFGAN~\cite{Yang2022MFGAN}\label{fig:real-3-e}]{
  \begin{minipage}{0.1405\linewidth}
    \centering
    \includegraphics[width=1\linewidth]{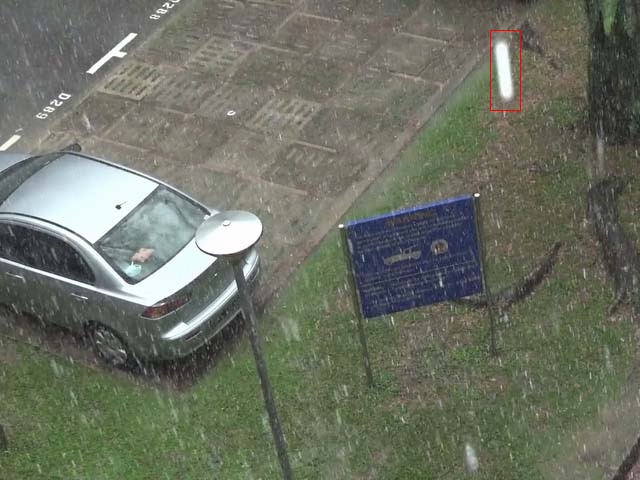}
    \hspace{-0mm}
    \includegraphics[width=1\linewidth]{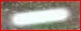}
     
    \centering
    \includegraphics[width=1\linewidth]{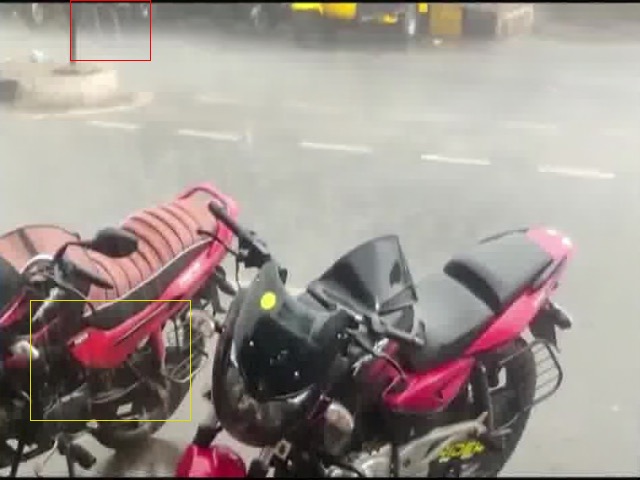}
    \hspace{-0mm}
    \includegraphics[width=0.490\linewidth]{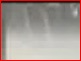}
    \hspace{-1.3mm}
     \includegraphics[width=0.490\linewidth]{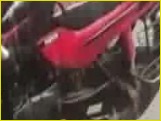}
    \end{minipage}
    }
    \hspace{-2.4mm}
    \subfloat[RainMamba~\cite{wu2024rainmamba}\label{fig:real-3-f}]{
  \begin{minipage}{0.1405\linewidth}
    \centering
    \includegraphics[width=1\linewidth]{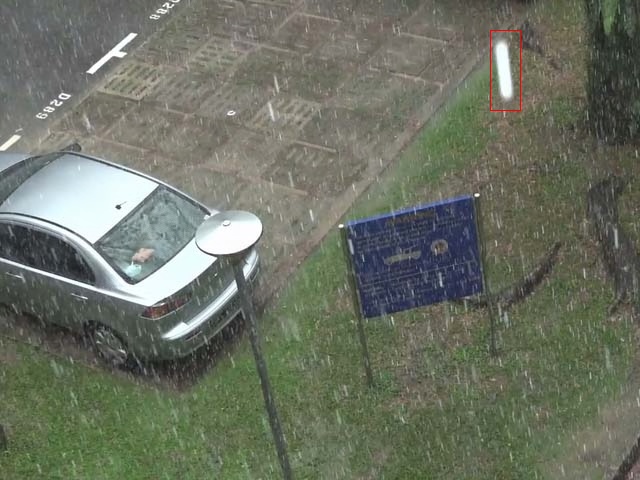}
    \hspace{-0mm}
    \includegraphics[width=1\linewidth]{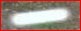}
     
    \centering
    \includegraphics[width=1\linewidth]{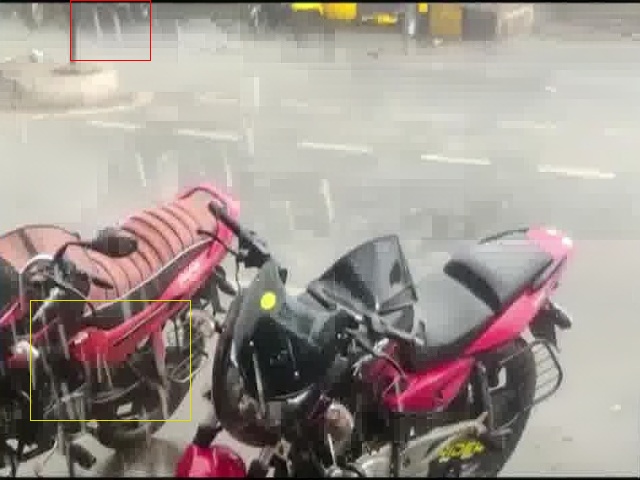}
    \hspace{-0mm}
    \includegraphics[width=0.490\linewidth]{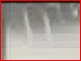}
    \hspace{-1.3mm}
     \includegraphics[width=0.490\linewidth]{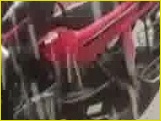}
    \end{minipage}
    }
    \hspace{-2.4mm}
    \subfloat[Ours\label{fig:real-3-g}]{
  \begin{minipage}{0.1405\linewidth}
    \centering
    \includegraphics[width=1\linewidth]{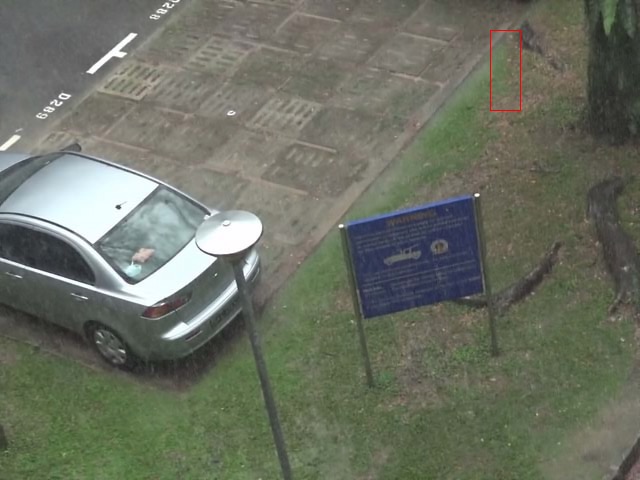}
    \hspace{-0mm}
    \includegraphics[width=1\linewidth]{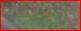}
     
    \centering
    \includegraphics[width=1\linewidth]{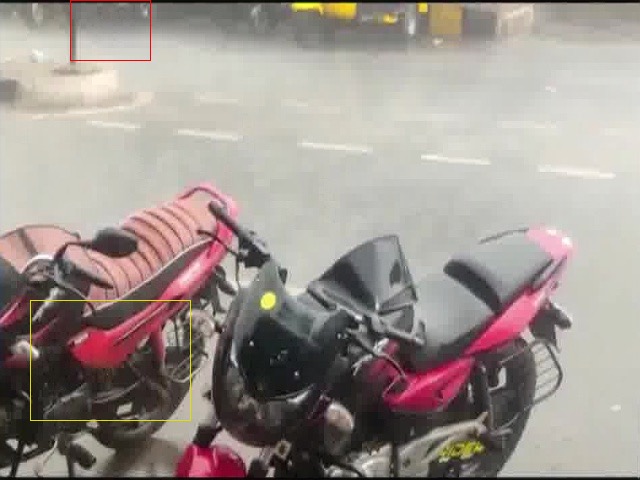}
    \hspace{-0mm}
    \includegraphics[width=0.490\linewidth]{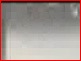}
    \hspace{-1.3mm}
     \includegraphics[width=0.490\linewidth]{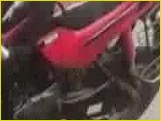}
    \end{minipage}
    }
    \vspace{-3mm}
  \caption{The qualitative comparison among the existing deraining methods on real-world rainy videos from \textit{NTURain}~\cite{Chen2018RobustCNN} and \textit{RVDT}. Please zoom in for better view.}
  \label{fig:real-3}
  \vspace{-4.5mm}
\end{figure*}

\vspace{-1mm}
\section{Experiment}
\vspace{-1mm}
\subsection{Experimental Settings}
\vspace{-1mm}
\noindent\textbf{Training Details.}
We train our model for 100 epochs and 50,000 iterations. 
The input data are cropped into patches of size $256\times 256$ with a sequence length of 7 frames. 
The initial learning rate is set to 0.0001, which is reduced by half after 50 epochs.
We utilize the optimizer Adam~\cite{Kingma2014Adam} with $\beta_1=0.9$ and $\beta_2=0.999$. 
The threshold $\theta$ and the slack number $\delta$ in Eq.~\ref{eq:stacking_patch_index_mask} are set to $80\%$ and 0.1, respectively. 
The loss weights are $\lambda_1=0.1$ and $\lambda_2=0.1$. 
The pre-trained optical flow estimator is LiteFlowNet3~\cite{hui2020liteflownet3}.
All experiments are conducted on Nvidia Tesla V100 32G GPU.

\noindent\textbf{Benchmarks.}
We evaluate our model on three widely used video deraining benchmarks: \textit{RainSynLight25}~\cite{Liu2018Erase}, \textit{RainSynComplex25}~\cite{Liu2018Erase}, and \textit{NTURain}~\cite{Chen2018RobustCNN}. 
To assess the potential of improving downstream high-level tasks, we also utilize our proposed RVDT benchmark.





\noindent\textbf{Baselines.}
We compare our method with state-of-the-art deraining methods, including 
FastDeRain~\cite{Jiang2019FastDerain}, DRSformer~\cite{Chen2023drsformer}, SpacCNN~\cite{Chen2018RobustCNN}, SLDNet~\cite{Yang2020Self}, J4RNet~\cite{yang2020joint}, S2VD~\cite{Yue2021Semi}, ESTINet~\cite{zhang2022enhanced}, MFGAN~\cite{Yang2022MFGAN}, and RainMamba~\cite{wu2024rainmamba}. 
Among them, FastDeRain~\cite{Jiang2019FastDerain} is a model-based method, and DRSformer~\cite{Chen2023drsformer} is a recent model for single image. 
The others are data-driven video-based methods.

\noindent\textbf{Metrics.}
In addition to the commonly used SSIM and PSNR metrics, we propose evaluating performance improvements in various object detection and trakcing tasks before and after the deraining process on selected rainy videos. 
Specifically, we use mAP (mean average precision) for object detection, and IDF1, MOTA, and MOTP~\cite{bernardin2008evaluating} for object tracking. 
We utilize two detection models: the image-based YOLO-v3~\cite{redmon2018yolov3} and the video-based MEGA~\cite{Chen2020MEGA}, along with GTR~\cite{zhou2022gtr} as the tracking method.


\begin{table}[tbp]
\footnotesize
\renewcommand{\arraystretch}{0.9}
\setlength{\tabcolsep}{1.5mm}
  \centering
  \caption{
  The object detection and tracking results on RVDT. 
  mAP$_1$ and mAP$_2$ denote the results of YOLO-v3~\cite{redmon2018yolov3} and MEGA~\cite{Chen2020MEGA}, respectively. 
  The tracking result is obtained by GTR~\cite{zhou2022gtr}.}
  \vspace{-3mm}
    \begin{tabular}{l|cc|ccc}
    \toprule
    Downstream Task & \multicolumn{2}{c|}{Object Detection} & \multicolumn{3}{c}{Object Tracking} \\
         Metrics & mAP$_1$$\uparrow$ & mAP$_2$$\uparrow$ & IDF1$\uparrow$ & MOTA$\uparrow$  & MOTP$\uparrow$ \\
    \midrule
    Rainy & 39.73 & 44.69 & 64.5  & 28.0  & 19.9 \\
    FastDeRain~\cite{Jiang2019FastDerain} & 40.45 & 42.90 & 66.5  & 29.8  & 20.9 \\
    SLDNet~\cite{Yang2020Self} & 41.59 & 42.06 & 66.7  & 30.7  & 20.4 \\
    DRSformer~\cite{Chen2023drsformer} & 39.10 & 44.01 & 67.0  & 28.8  & 20.0 \\
    S2VD~\cite{Yue2021Semi}  & 41.37 & 43.73 & 68.1  & 29.4  & 20.4 \\
    ESTINet~\cite{zhang2022enhanced} & 40.71 & 43.05 & 67.6  & 29.7  & 19.9 \\
    MFGAN~\cite{Yang2022MFGAN} & 40.83 & 43.97 & 67.1  & 28.6  & 20.1 \\
    RainMamba~\cite{wu2024rainmamba} & 40.67 & 44.22 & 67.2 & 29.5 & 20.3 \\
    Ours  & \textbf{42.51} & \textbf{46.27} & \textbf{68.3} & \textbf{31.8} & \textbf{21.2} \\
    \bottomrule
    \end{tabular}%
  \label{tab:rrodt_result}%
  \vspace{-3mm}
\end{table}%

\begin{figure*}
  \centering
  \hspace{-2mm}
  \subfloat[Rainy\label{fig:rvdt-a}]{
  \begin{minipage}{0.1405\linewidth}
    \centering
    \includegraphics[width=1\linewidth]{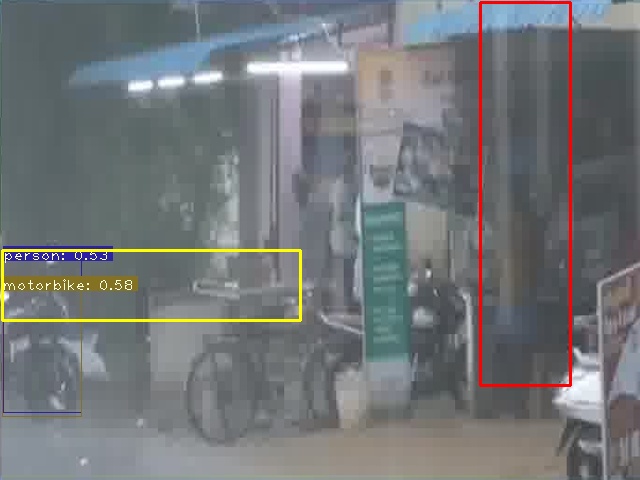}
    \hspace{-0mm}
    \includegraphics[width=1\linewidth]{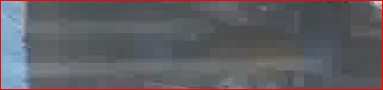}
    \hspace{-1.3mm}
    \includegraphics[width=1\linewidth]{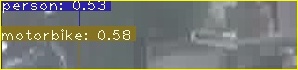}
    
    \vspace{1mm}
    \centering
    \includegraphics[width=1\linewidth]{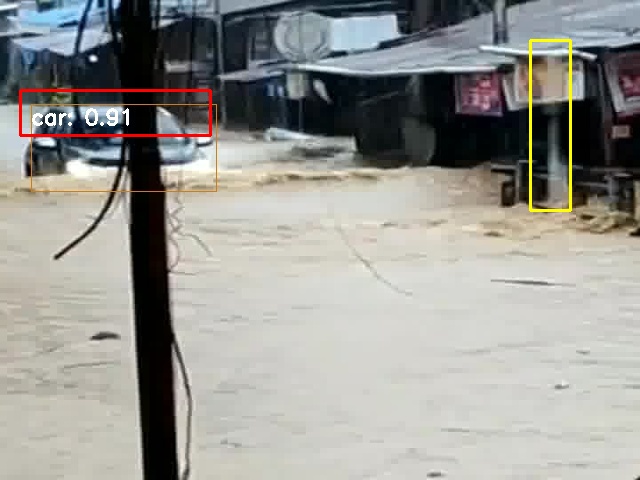}
    \hspace{-0mm}
    \includegraphics[width=1\linewidth]{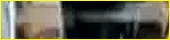}
    \hspace{-1.3mm}
    \includegraphics[width=1\linewidth]{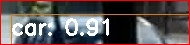}
    
    \vspace{1mm}
    \centering
    \includegraphics[width=1\linewidth]{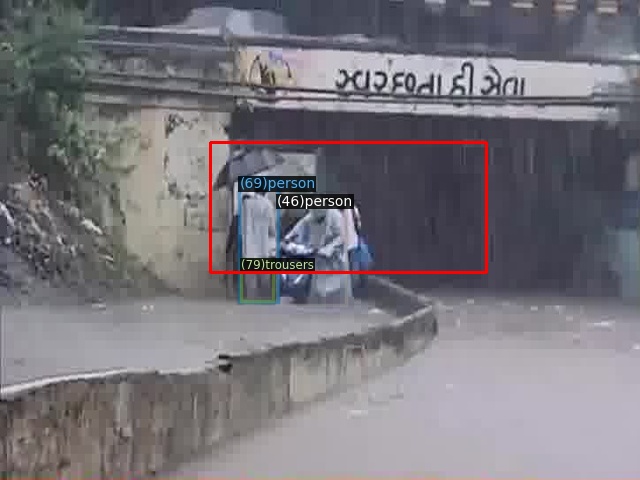}
    \hspace{-0mm}
    \includegraphics[width=1\linewidth]{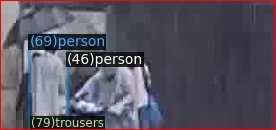}
    \end{minipage}
    }
    \hspace{-2.4mm}
    \subfloat[FastDeRain~\cite{Jiang2019FastDerain}\label{fig:rvdt-b}]{
  \begin{minipage}{0.1405\linewidth}
    \centering
    \includegraphics[width=1\linewidth]{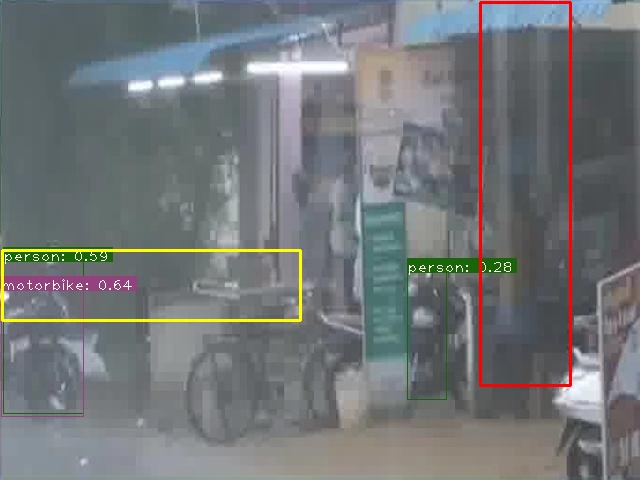}
    \hspace{-0mm}
    \includegraphics[width=1\linewidth]{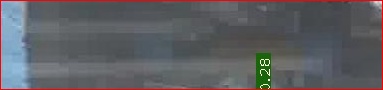}
    \hspace{-1.3mm}
    \includegraphics[width=1\linewidth]{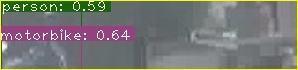}
     
    \vspace{1mm}
    \centering
    \includegraphics[width=1\linewidth]{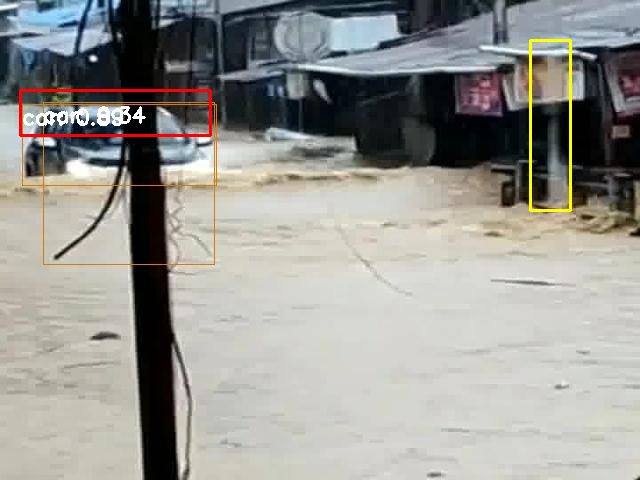}
    \hspace{-0mm}
    \includegraphics[width=1\linewidth]{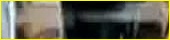}
    \hspace{-1.3mm}
    \includegraphics[width=1\linewidth]{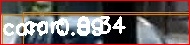}
    
    \vspace{1mm}
    \centering
    \includegraphics[width=1\linewidth]{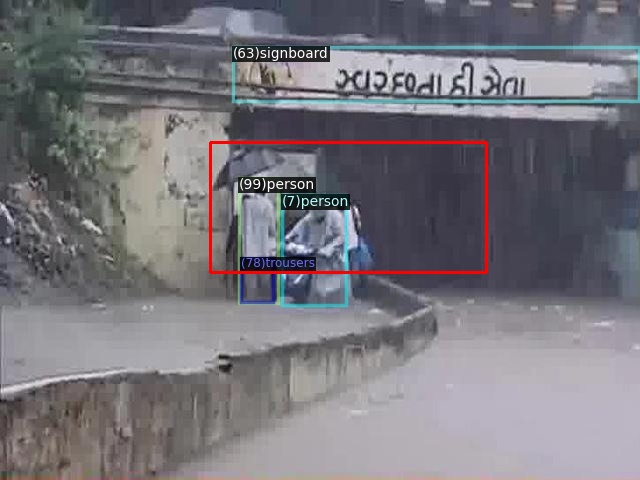}
    \hspace{-0mm}
    \includegraphics[width=1\linewidth]{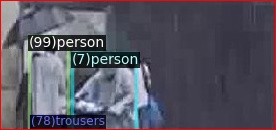}
    \end{minipage}
    }
    \hspace{-2.4mm}
    \subfloat[DRSformer.~\cite{Chen2023drsformer}\label{fig:rvdt-c}]{
  \begin{minipage}{0.1405\linewidth}
    \centering
    \includegraphics[width=1\linewidth]{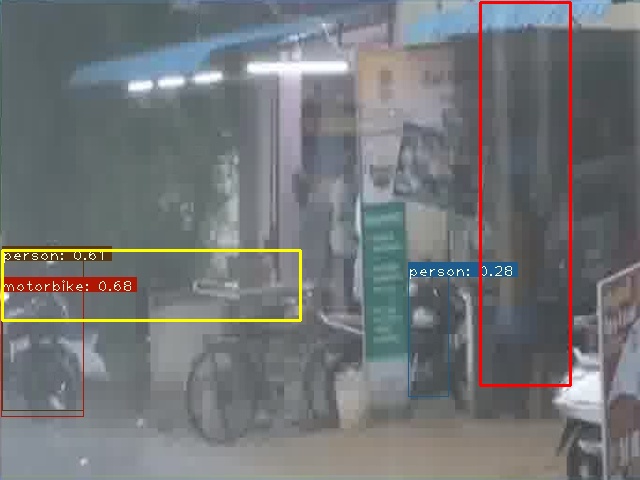}
    \hspace{-0mm}
    \includegraphics[width=1\linewidth]{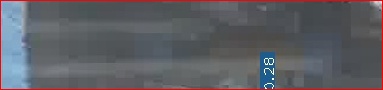}
    \hspace{-0mm}
    \includegraphics[width=1\linewidth]{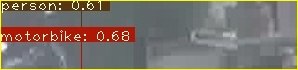}
     
    \vspace{1mm}
    \centering
    \includegraphics[width=1\linewidth]{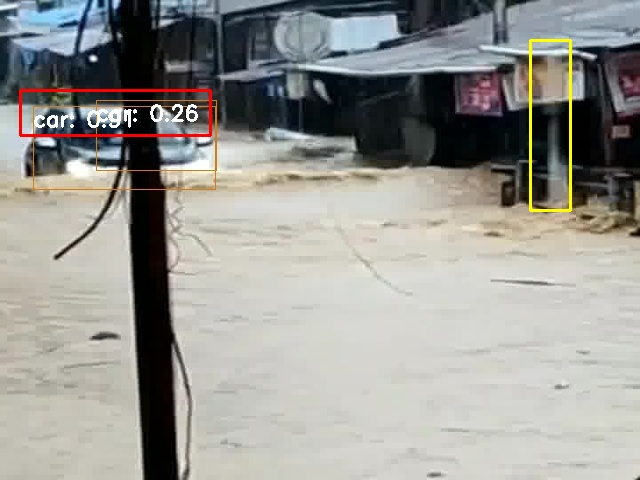}
    \hspace{-0mm}
    \includegraphics[width=1\linewidth]{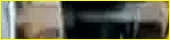}
    \hspace{-0mm}
    \includegraphics[width=1\linewidth]{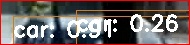}
    
    \vspace{1mm}
    \centering
    \includegraphics[width=1\linewidth]{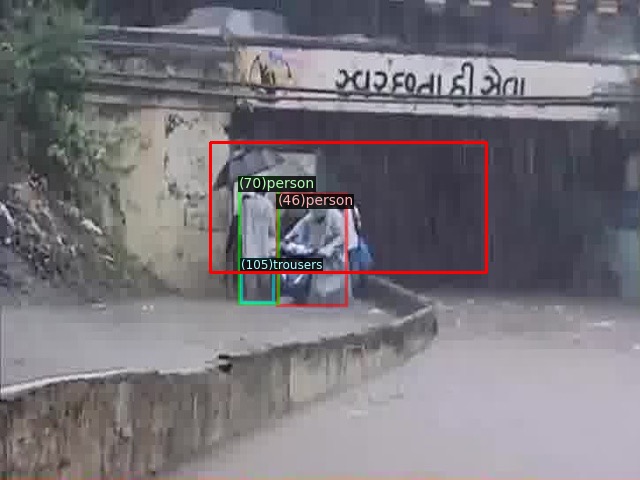}
    \hspace{-0mm}
    \includegraphics[width=1\linewidth]{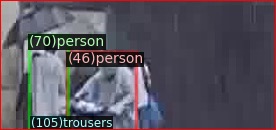}
    \end{minipage}
    }
    \hspace{-2.4mm}
    \subfloat[ESTINet~\cite{zhang2022enhanced}\label{fig:rvdt-d}]{
  \begin{minipage}{0.1405\linewidth}
    \centering
    \includegraphics[width=1\linewidth]{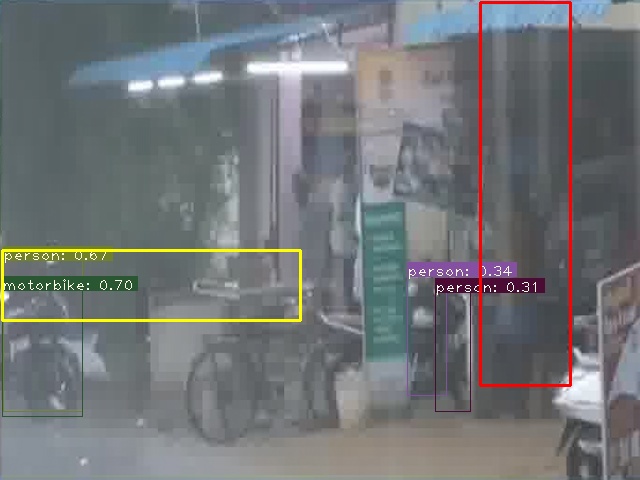}
    \hspace{-0mm}
    \includegraphics[width=1\linewidth]{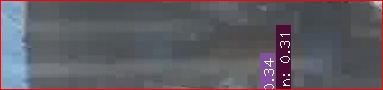}
    \hspace{-1.3mm}
    \includegraphics[width=1\linewidth]{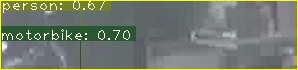}
     
    \vspace{1mm}
    \centering
    \includegraphics[width=1\linewidth]{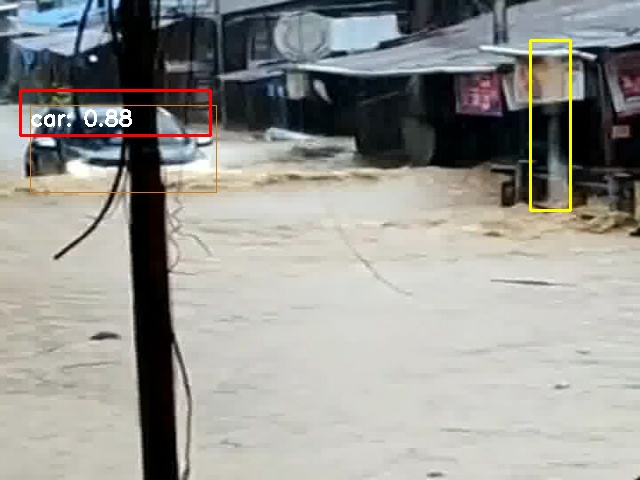}
    \hspace{-0mm}
    \includegraphics[width=1\linewidth]{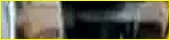}
    \hspace{-1.3mm}
    \includegraphics[width=1\linewidth]{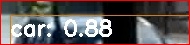}
    
    \vspace{1mm}
    \centering
    \includegraphics[width=1\linewidth]{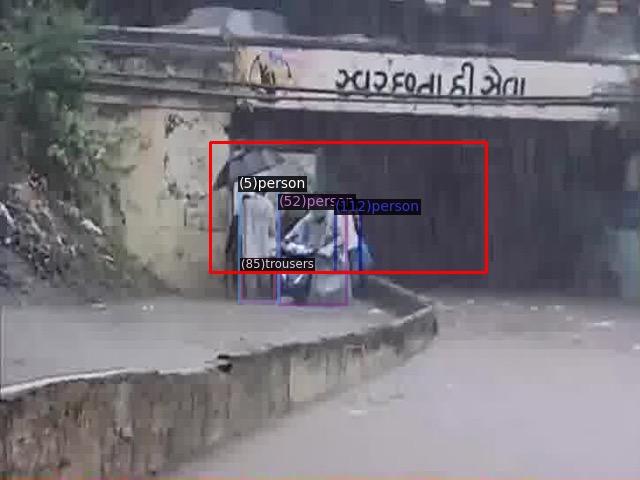}
    \hspace{-0mm}
    \includegraphics[width=1\linewidth]{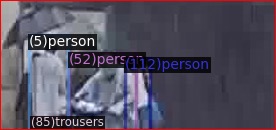}
    \end{minipage}
    }
    \hspace{-2.4mm}
    \subfloat[MFGAN~\cite{Yang2022MFGAN}\label{fig:rvdt-e}]{
  \begin{minipage}{0.1405\linewidth}
    \centering
    \includegraphics[width=1\linewidth]{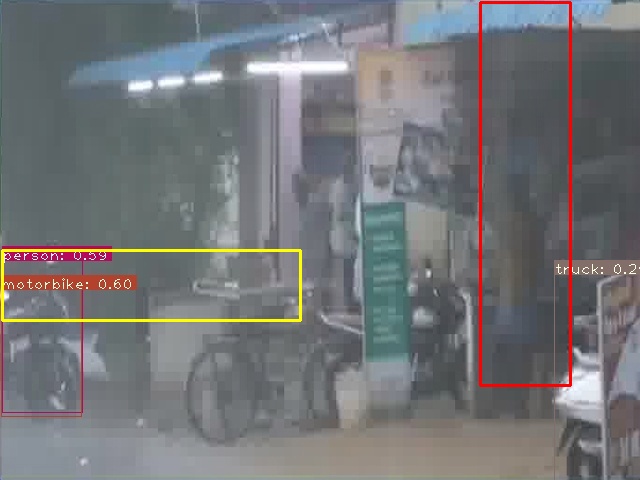}
    \hspace{-0mm}
    \includegraphics[width=1\linewidth]{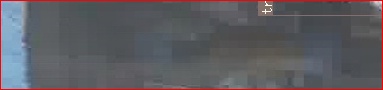}
    \hspace{-1.3mm}
    \includegraphics[width=1\linewidth]{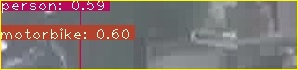}
     
    \vspace{1mm}
    \centering
    \includegraphics[width=1\linewidth]{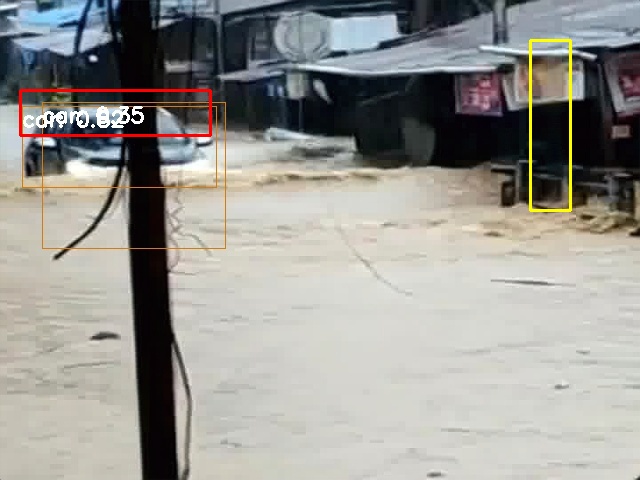}
    \hspace{-0mm}
    \includegraphics[width=1\linewidth]{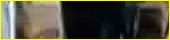}
    \hspace{-1.3mm}
    \includegraphics[width=1\linewidth]{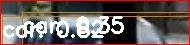}
    
    \vspace{1mm}
    \centering
    \includegraphics[width=1\linewidth]{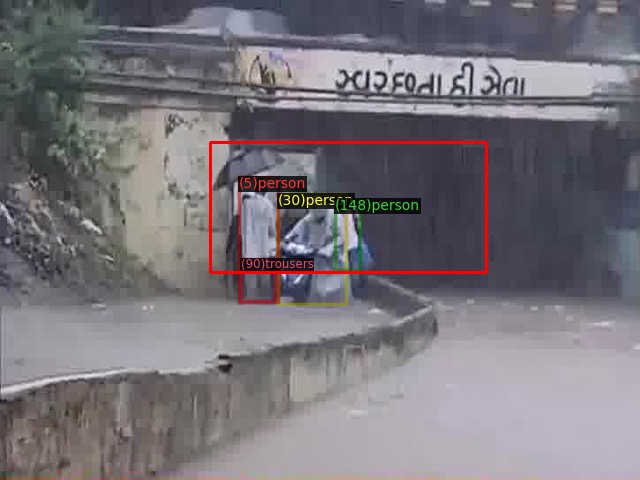}
    \hspace{-0mm}
    \includegraphics[width=1\linewidth]{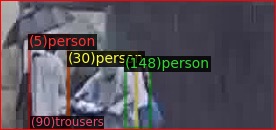}
    \end{minipage}
    }
    \hspace{-2.4mm}
    \subfloat[RainMamba~\cite{wu2024rainmamba}\label{fig:rvdt-f}]{
  \begin{minipage}{0.1405\linewidth}
    \centering
    \includegraphics[width=1\linewidth]{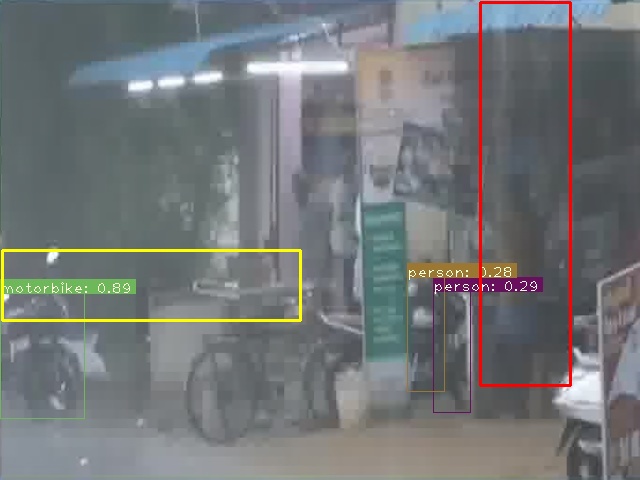}
    \hspace{-0mm}
    \includegraphics[width=1\linewidth]{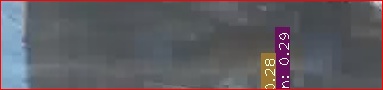}
    \hspace{-1.3mm}
    \includegraphics[width=1\linewidth]{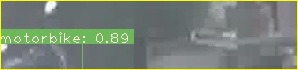}
     
    \vspace{1mm}
    \centering
    \includegraphics[width=1\linewidth]{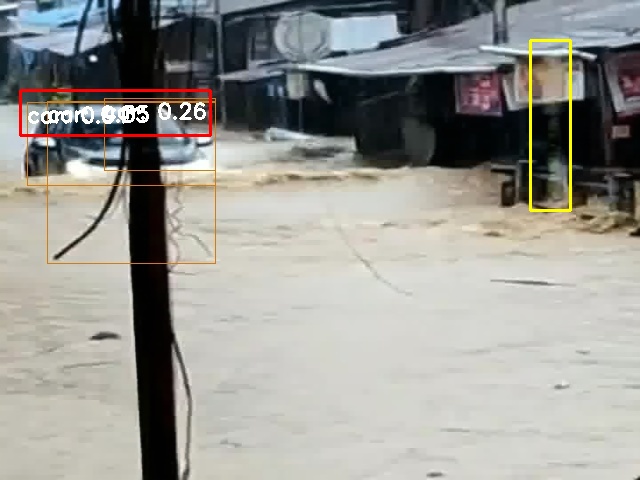}
    \hspace{-0mm}
    \includegraphics[width=1\linewidth]{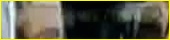}
    \hspace{-1.3mm}
    \includegraphics[width=1\linewidth]{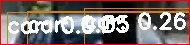}
    
    \vspace{1mm}
    \centering
    \includegraphics[width=1\linewidth]{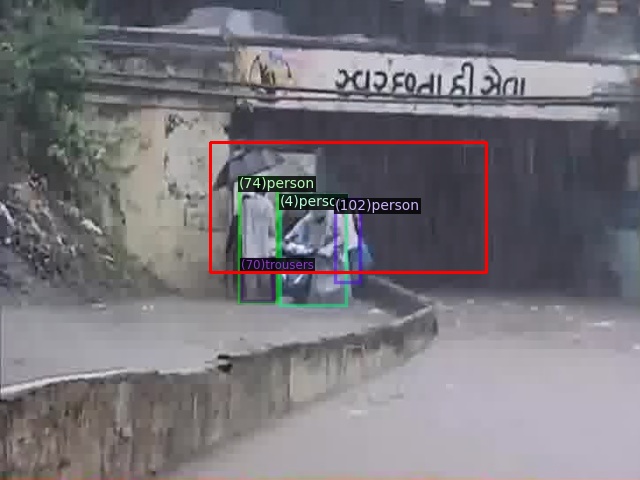}
    \hspace{-0mm}
    \includegraphics[width=1\linewidth]{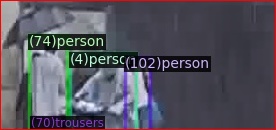}
    \end{minipage}
    }
    \hspace{-2.4mm}
    \subfloat[Ours\label{fig:rvdt-g}]{
  \begin{minipage}{0.1405\linewidth}
    \centering
    \includegraphics[width=1\linewidth]{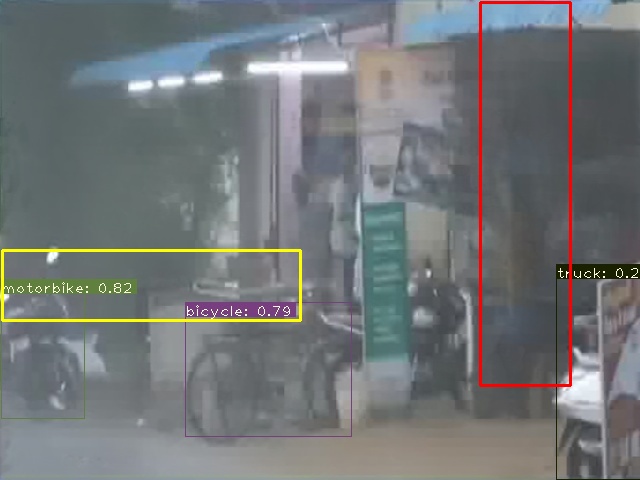}
    \hspace{-0mm}
    \includegraphics[width=1\linewidth]{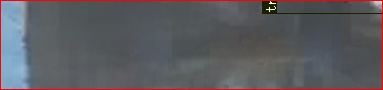}
    \hspace{-1.3mm}
    \includegraphics[width=1\linewidth]{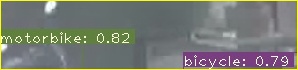}
     
    \vspace{1mm}
    \centering
    \includegraphics[width=1\linewidth]{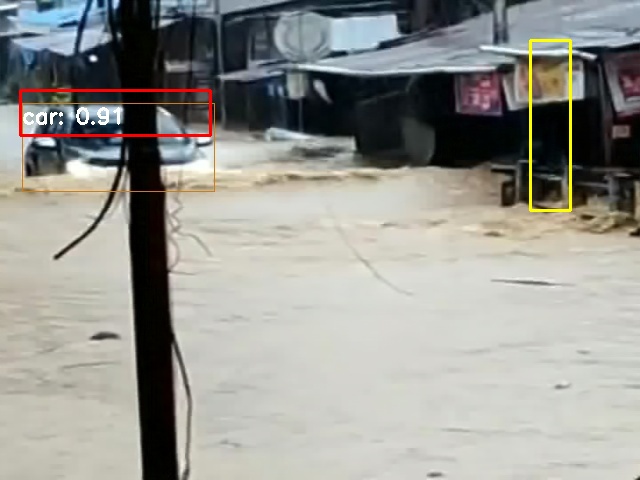}
    \hspace{-0mm}
    \includegraphics[width=1\linewidth]{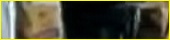}
    \hspace{-1.3mm}
    \includegraphics[width=1\linewidth]{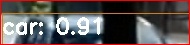}

    \vspace{1mm}
    \centering
    \includegraphics[width=1\linewidth]{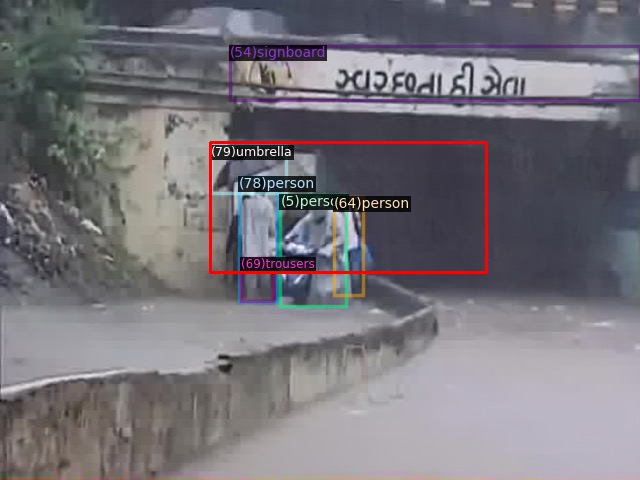}
    \hspace{-0mm}
    \includegraphics[width=1\linewidth]{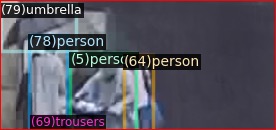}
    \end{minipage}
    }
    \vspace{-3mm}
  \caption{Rainy-scene object detection and tracking results on RVDT after video deraining.
  Three results are obtained using YOLO-v3~\cite{redmon2018yolov3}, MEGA~\cite{Chen2020MEGA}, and GTR~\cite{zhou2022gtr}, respectively. 
  Please zoom in for better view.}
  \label{fig:rvdt}
  \vspace{-4mm}
\end{figure*}

\begin{figure}[tb]
  \centering
  \subfloat[Rainy]{\label{fig:ablation_stacking_loss-rainy1}\centering\includegraphics[width=0.331\linewidth]{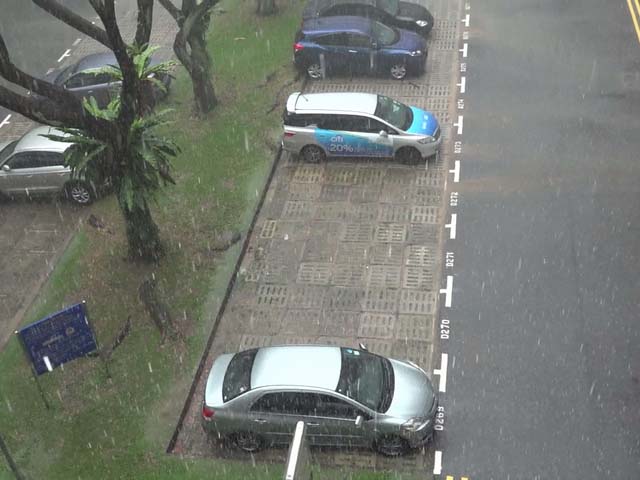}}
  \hfill
  \subfloat[$\theta=40\%$]{\label{fig:ablation_stacking_loss-flow1}\centering\includegraphics[width=0.331\linewidth]{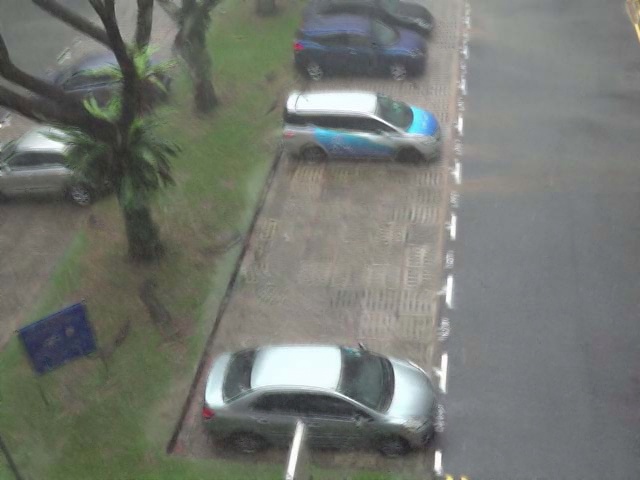}}
  \hfill
  \subfloat[$\theta=60\%$]{\label{fig:ablation_stacking_loss-warp1}\centering\includegraphics[width=0.331\linewidth]{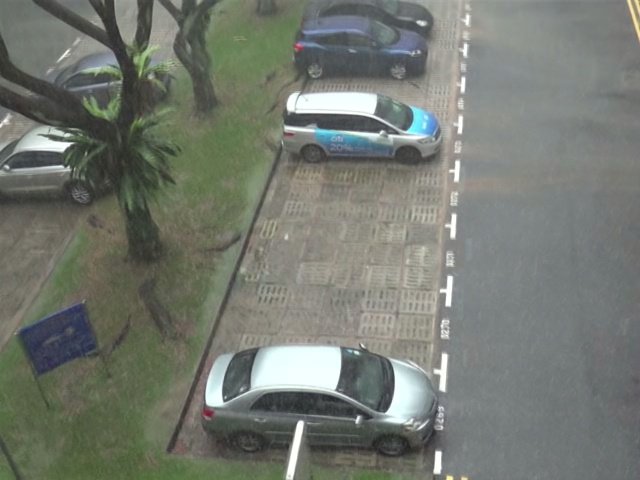}}

  \centering
  \subfloat[$\theta=80\%$]{\label{fig:ablation_stacking_loss-rainy2}\centering\includegraphics[width=0.331\linewidth]{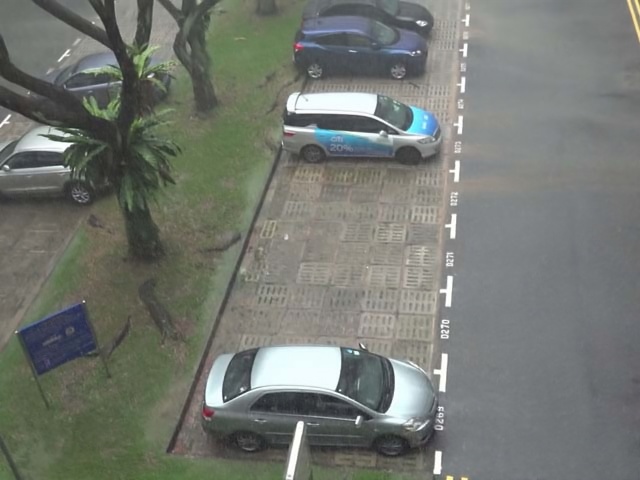}}
  \hfill
  \subfloat[$\theta=100\%$]{\label{fig:ablation_stacking_loss-flow2}\centering\includegraphics[width=0.331\linewidth]{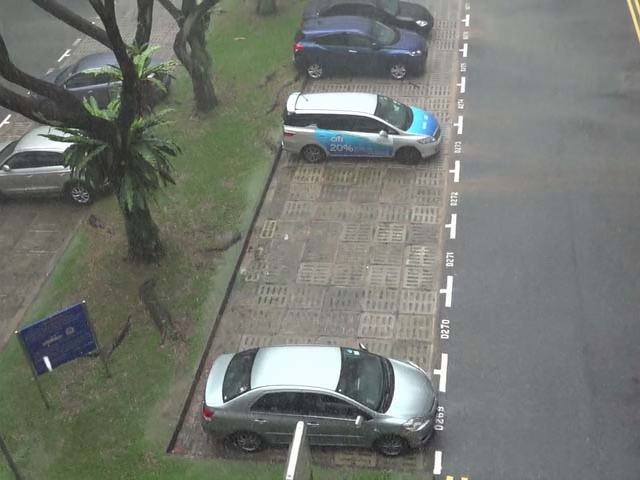}}
  \hfill
  \subfloat[No stacking loss]{\label{fig:ablation_stacking_loss-warp2}\centering\includegraphics[width=0.331\linewidth]{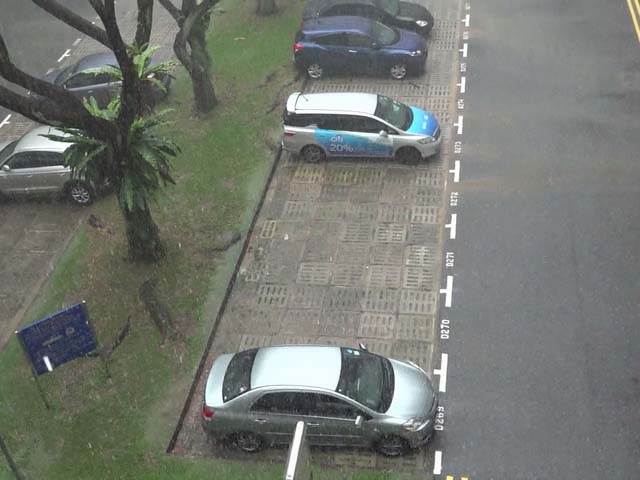}}
\setlength{\abovecaptionskip}{1pt} 
\setlength{\belowcaptionskip}{-1pt}
  \caption{The qualitative ablation of the stacking loss.}
  \label{fig:ablation_stacking_loss}
  \vspace{-3mm}
\end{figure}
\begin{table}[tbp]
\footnotesize
\renewcommand{\arraystretch}{0.9}
  \centering
  \caption{The quantitative results of ablation analysis.}
  \vspace{-3mm}
    \begin{tabular}{lcccc}
    \toprule
          & \multicolumn{1}{c}{Dual-branch U-Net} & \multicolumn{1}{c}{w/o S3ML} & \multicolumn{1}{c}{w/o TSML} & \multicolumn{1}{c}{Ours} \\
    \midrule
    PSNR  & 38.54 & 39.02 & 39.24 & 39.74 \\
    SSIM  & 0.9733 & 0.9754 & 0.9763 & 0.9791 \\
    \midrule
          & \multicolumn{1}{c}{Strategy 1} & \multicolumn{1}{c}{Strategy 2} & \multicolumn{1}{c}{Strategy 3} & \multicolumn{1}{c}{Ours} \\
    \midrule
    PSNR  & 39.65 & 39.03 & 38.51 & 39.74 \\
    SSIM  & 0.9758 & 0.9754 & 0.9734 & 0.9791 \\
    \midrule
          & \multicolumn{1}{c}{w/o DSF} & \multicolumn{1}{c}{median} & \multicolumn{1}{c}{mean} & \multicolumn{1}{c}{Ours} \\
    \midrule
    PSNR  & 39.21 & 39.09 & 39.31 & 39.74 \\
    SSIM  & 0.9755 & 0.9749 & 0.9761 & 0.9791 \\
    \bottomrule
    \end{tabular}%
  \label{tab:ablation}%
  \vspace{-4mm}
\end{table}%


\vspace{-0.5mm}
\subsection{Video Deraining Result}
\vspace{-0.5mm}

\noindent\textbf{Synthetic Rain Removal.}
We present the quantitative comparisons on three synthetic datasets
in Tab.~\ref{tab:rainsyn}. 
Our method achieves the highest PSNR and SSIM values across all the benchmarks. 
For a visual quality comparison, we provide examples in Fig.~\ref{fig:realra3}, illustrating that only our method successfully removes all rain streaks.


\vspace{0.5mm}
\noindent\textbf{Real-World Rain Removal.}
We then assess the effectiveness of the existing deraining methods for real-world video deraining. 
A group of visual comparisons are presented in Fig.~\ref{fig:real-3}. 
As demonstrated, only our method successfully removes all rain streaks, regardless of their scale. 

\vspace{0.5mm}
\noindent\textbf{Complexity Analysis}
We compare our VDMamba with other deraining methods in terms of time and space complexities in Fig.~\ref{fig:complexity}. 
``Ours-Single'' refers to the lightweight version of our VDMamba, which includes only the spatial branch. 
Our method achieves the best overall balance between deraining performance and the complexities.

\vspace{-0.5mm}
\subsection{Video Deraining for Real-World Tasks}
\vspace{-0.5mm}

\noindent\textbf{Rainy Object Detection.}
We conduct experiments of real-world rainy object detection using our RVDT. 
The quantitative results are presented in Tab.~\ref{tab:rrodt_result}, while visual comparisons are shown in the first two examples of Fig.~\ref{fig:rvdt}.
Only our method successfully removes the rain streaks, while simultaneously improving detection accuracy. 
%


\vspace{0.5mm}
\noindent\textbf{Rainy Object Tracking.}
We also evaluate the tracking performance on the videos restored by the deraining methods. 
The results are presented in the right half of Tab.~\ref{tab:rrodt_result}. 
Our method demonstrates the most significant improvement compared to direct tracking on the rainy videos. 
Additionally, a visual comparison is provided in the final example of Fig.~\ref{fig:rvdt}.
It is evident that only our method effectively removes all rain streaks and enables the tracking method to correctly identify both the object ``umbrella'' and the ``person''..

\vspace{-0.5mm}
\subsection{Ablation Studies}
\vspace{-0.5mm}

To validate the necessity of the components we introduced, we conduct four groups of ablation studies.

\vspace{0.5mm}
\noindent\textbf{Network Structure.} 
We design three additional model variants: (1) replacing both S3ML and TSML with residual blocks to form a pure dual-branch U-Net~\cite{Ronneberger2015UNetCN}, (2) replacing S3ML alone, and (3) replacing TSML alone. 
The results in the first row of Tab.~\ref{tab:ablation} shows that both S3ML and TSML contribute to improving the deraining performance.

\vspace{0.5mm}
\noindent\textbf{Optical Flow Estimation.}
We design three optical flow estimation strategies. Strategy 1 directly uses the pre-trained model to estimate flows for multiple frame pairs. 
Strategy 2 and 3 involve conducting the transfer on rainy frames and clean frames, respectively.
The results are presented in the second row of Tab.~\ref{tab:ablation}. 
Though Strategy 1 achieves a comparable result, its speed is significantly slower.

\vspace{0.5mm}
\noindent\textbf{Dynamic Stacking Filter.}
Apart from the proposed dynamic stacking filter, we evaluate three additional fusion strategies: removing the filter, using the median filter, and using the mean filter. 
The results are shown in the last row of Tab.~\ref{tab:ablation}. 
Our method demonstrates a performance improvement over 
the other strategies.

\vspace{0.5mm}
\noindent\textbf{Threshold of Stacking Loss.}
We experiment with different values for the threshold $\theta$ to generate the mask index for learning sub-patches in Eq.~\ref{eq:stacking_patch_index_mask}. 
The resulting deraining examples are shown in Fig.~\ref{fig:ablation_stacking_loss}.
We observe that the introduction of the stacking loss enhances the model's capability for real-world deraining. 
However, a too slack threshold, such as $\theta=40\%$, causes the network to learn from many mismatched sub-patches, resulting in blurry outputs. 
As a threshold range from $60\%$ to $90\%$ provides satisfactory results, we select $80\%$ as the default value.


\section{Conclusion}
In this work, we propose a dual-branch spatio-temporal state-space model for video restoration in rainy conditions, addressing the gap between synthetic and real-world rain effects. 
By integrating spatial and temporal feature extraction, dynamic stacking filters, and a median stacking loss, our method effectively enhances rain streak removal. 
Extensive evaluations on synthetic and real-world rainy videos demonstrate its superior performance in visual quality, efficiency, and our proposed downstream vision benchmarks.

\clearpage
\section*{Acknowledgment}
This work was supported in part by the National Key R\&D Program of China (Grant No.2022ZD0119200), in part by National Natural Science Foundation of China (No. 62322216, 62025604, 62172409), and in part by Shenzhen Science and Technology Program (Grant No. JCYJ20220818102012025, KQTD20221101093559018, RCYX20221008092849068).
{
    \small
    \bibliographystyle{ieeenat_fullname}
    \bibliography{main}

\begin{thebibliography}{74}
\providecommand{\natexlab}[1]{#1}
\providecommand{\url}[1]{\texttt{#1}}
\expandafter\ifx\csname urlstyle\endcsname\relax
  \providecommand{\doi}[1]{doi: #1}\else
  \providecommand{\doi}{doi: \begingroup \urlstyle{rm}\Url}\fi

\bibitem[Aloysius and Geetha(2017)]{aloysius2017review}
Neena Aloysius and M Geetha.
\newblock A review on deep convolutional neural networks.
\newblock In \emph{ICCSP}, 2017.

\bibitem[Barnum et~al.(2007)Barnum, Kanade, and Narasimhan]{Barnum07spatio-temporalfrequency}
Peter Barnum, Takeo Kanade, and Srinivasa~G Narasimhan.
\newblock Spatio-temporal frequency analysis for removing rain and snow from videos.
\newblock In \emph{PACV}, 2007.

\bibitem[Barnum et~al.(2010)Barnum, Narasimhan, and Kanade]{Barnum2010Analysis}
Peter Barnum, Srinivasa Narasimhan, and Takeo Kanade.
\newblock Analysis of rain and snow in frequency space.
\newblock \emph{IJCV}, 2010.

\bibitem[Bernardin and Stiefelhagen(2008)]{bernardin2008evaluating}
Keni Bernardin and Rainer Stiefelhagen.
\newblock Evaluating multiple object tracking performance: the clear mot metrics.
\newblock \emph{JIVP}, 2008.

\bibitem[Bossu et~al.(2011)Bossu, Hauti\`ere, and Tarel]{Bossu2011Rain}
J. Bossu, N. Hauti\`ere, and J.-P. Tarel.
\newblock Rain or snow detection in image sequences through use of a histogram of orientation of streaks.
\newblock \emph{IJCV}, 2011.

\bibitem[Chen et~al.(2014)Chen, Chen, and Kang]{chen2014visual}
Duan-Yu Chen, Chien-Cheng Chen, and Li-Wei Kang.
\newblock Visual depth guided color image rain streaks removal using sparse coding.
\newblock \emph{IEEE TCSVT}, 2014.

\bibitem[Chen et~al.(2018)Chen, Tan, Hou, Chau, and Li]{Chen2018RobustCNN}
Jie Chen, Cheen-Hau Tan, Junhui Hou, Lap-Pui Chau, and He Li.
\newblock Robust video content alignment and compensation for rain removal in a cnn framework.
\newblock In \emph{CVPR}, 2018.

\bibitem[Chen et~al.(2023)Chen, Li, Li, and Pan]{Chen2023drsformer}
Xiang Chen, Hao Li, Mingqiang Li, and Jinshan Pan.
\newblock Learning a sparse transformer network for effective image deraining.
\newblock In \emph{CVPR}, 2023.

\bibitem[Chen et~al.(2020)Chen, Cao, Hu, and Wang]{Chen2020MEGA}
Yihong Chen, Yue Cao, Han Hu, and Liwei Wang.
\newblock Memory enhanced global-local aggregation for video object detection.
\newblock In \emph{CVPR}, 2020.

\bibitem[Chen and Hsu(2013)]{Chen2013LowRankModel}
Yi-Lei Chen and Chiou-Ting Hsu.
\newblock A generalized low-rank appearance model for spatio-temporally correlated rain streaks.
\newblock In \emph{ICCV}, 2013.

\bibitem[Elfwing et~al.(2018)Elfwing, Uchibe, and Doya]{elfwing2018sigmoid}
Stefan Elfwing, Eiji Uchibe, and Kenji Doya.
\newblock Sigmoid-weighted linear units for neural network function approximation in reinforcement learning.
\newblock \emph{Neural Netw.}, 2018.

\bibitem[Fu et~al.(2017{\natexlab{a}})Fu, Huang, Ding, Liao, and Paisley]{fu2017clearing}
Xueyang Fu, Jiabin Huang, Xinghao Ding, Yinghao Liao, and John Paisley.
\newblock Clearing the skies: A deep network architecture for single-image rain removal.
\newblock \emph{IEEE TIP}, 2017{\natexlab{a}}.

\bibitem[Fu et~al.(2017{\natexlab{b}})Fu, Huang, Zeng, Huang, Ding, and Paisley]{fu2017removing}
Xueyang Fu, Jiabin Huang, Delu Zeng, Yue Huang, Xinghao Ding, and John Paisley.
\newblock Removing rain from single images via a deep detail network.
\newblock In \emph{CVPR}, 2017{\natexlab{b}}.

\bibitem[Garg and Nayar(2004)]{Garg2004DetectionAR}
K. Garg and S. Nayar.
\newblock Detection and removal of rain from videos.
\newblock In \emph{CVPR}, 2004.

\bibitem[Garg and Nayar(2005)]{Garg2005ICCV}
Kshitiz Garg and Shree~K. Nayar.
\newblock When does a camera see rain?
\newblock In \emph{ICCV}, 2005.

\bibitem[Garg and Nayar(2007)]{garg2007visionRain}
Kshitiz Garg and Shree~K. Nayar.
\newblock Vision and rain.
\newblock \emph{IJCV}, 2007.

\bibitem[Gu and Dao(2023)]{gu2023mamba}
Albert Gu and Tri Dao.
\newblock Mamba: Linear-time sequence modeling with selective state spaces.
\newblock \emph{arXiv preprint arXiv:2312.00752}, 2023.

\bibitem[Gu et~al.(2021)Gu, Goel, and R{\'e}]{gu2021mamba}
Albert Gu, Karan Goel, and Christopher R{\'e}.
\newblock Efficiently modeling long sequences with structured state spaces.
\newblock \emph{arXiv preprint arXiv:2111.00396}, 2021.

\bibitem[Guo et~al.(2024)Guo, Li, Dai, Ouyang, Ren, and Xia]{guo2024mambair}
Hang Guo, Jinmin Li, Tao Dai, Zhihao Ouyang, Xudong Ren, and Shu-Tao Xia.
\newblock Mambair: A simple baseline for image restoration with state-space model.
\newblock \emph{arXiv preprint arXiv:2402.15648}, 2024.

\bibitem[Hassaballah et~al.(2020)Hassaballah, Kenk, Muhammad, and Minaee]{hassaballah2020vehicle}
Mahmoud Hassaballah, Mourad~A Kenk, Khan Muhammad, and Shervin Minaee.
\newblock Vehicle detection and tracking in adverse weather using a deep learning framework.
\newblock \emph{IEEE TITS}, 2020.

\bibitem[Hatamizadeh and Kautz(2024)]{hatamizadeh2024mambavision}
Ali Hatamizadeh and Jan Kautz.
\newblock Mambavision: A hybrid mamba-transformer vision backbone.
\newblock \emph{arXiv preprint arXiv:2407.08083}, 2024.

\bibitem[Hinton et~al.(2015)Hinton, Vinyals, and Dean]{Hinton2015DistillingTK}
Geoffrey~E. Hinton, Oriol Vinyals, and Jeffrey Dean.
\newblock Distilling the knowledge in a neural network.
\newblock \emph{ArXiv}, abs/1503.02531, 2015.

\bibitem[Hnewa and Radha(2020)]{hnewa2020object}
Mazin Hnewa and Hayder Radha.
\newblock Object detection under rainy conditions for autonomous vehicles: A review of state-of-the-art and emerging techniques.
\newblock \emph{IEEE SPM}, 2020.

\bibitem[Hui and Loy(2020)]{hui2020liteflownet3}
Tak-Wai Hui and Chen~Change Loy.
\newblock Liteflownet3: Resolving correspondence ambiguity for more accurate optical flow estimation.
\newblock In \emph{ECCV}, 2020.

\bibitem[Jiang et~al.(2020)Jiang, Wang, Yi, Chen, Huang, Luo, Ma, and Jiang]{Jiang2020MultiScale}
Kui Jiang, Zhongyuan Wang, Peng Yi, Chen Chen, Baojin Huang, Yimin Luo, Jiayi Ma, and Junjun Jiang.
\newblock Multi-scale progressive fusion network for single image deraining.
\newblock In \emph{CVPR}, 2020.

\bibitem[Jiang et~al.(2023)Jiang, Liu, Wang, Zhong, Jiang, and Lin]{jiang2023dawn}
Kui Jiang, Wenxuan Liu, Zheng Wang, Xian Zhong, Junjun Jiang, and Chia-Wen Lin.
\newblock Dawn: Direction-aware attention wavelet network for image deraining.
\newblock In \emph{ACM MM}, 2023.

\bibitem[Jiang et~al.(2017)Jiang, Huang, Zhao, Deng, and Wang]{Jiang2017NovelTensorBased}
Tai-Xiang Jiang, Ting-Zhu Huang, Xi-Le Zhao, Liang-Jian Deng, and Yao Wang.
\newblock A novel tensor-based video rain streaks removal approach via utilizing discriminatively intrinsic priors.
\newblock In \emph{CVPR}, 2017.

\bibitem[Jiang et~al.(2019)Jiang, Huang, Zhao, Deng, and Wang]{Jiang2019FastDerain}
Tai-Xiang Jiang, Ting-Zhu Huang, Xi-Le Zhao, Liang-Jian Deng, and Yao Wang.
\newblock Fastderain: A novel video rain streak removal method using directional gradient priors.
\newblock \emph{IEEE TIP}, 2019.

\bibitem[Kang et~al.(2012)Kang, Lin, and Fu]{kang2012automatic}
Li-Wei Kang, Chia-Wen Lin, and Yu-Hsiang Fu.
\newblock Automatic single-image-based rain streaks removal via image decomposition.
\newblock \emph{IEEE TIP}, 2012.

\bibitem[Kim et~al.(2013)Kim, Lee, Sim, and Kim]{kim2013single}
Jin-Hwan Kim, Chul Lee, Jae-Young Sim, and Chang-Su Kim.
\newblock Single-image deraining using an adaptive nonlocal means filter.
\newblock In \emph{ICIP}, 2013.

\bibitem[Kim et~al.(2015)Kim, Sim, and Kim]{Kim2015VideoDeraining}
Jin-Hwan Kim, Jae-Young Sim, and Chang-Su Kim.
\newblock Video deraining and desnowing using temporal correlation and low-rank matrix completion.
\newblock \emph{IEEE TIP}, 2015.

\bibitem[Kingma and Ba(2015)]{Kingma2014Adam}
Diederik~P Kingma and Jimmy Ba.
\newblock Adam: A method for stochastic optimization.
\newblock In \emph{ICLR}, 2015.

\bibitem[Li et~al.(2024)Li, Liu, Fu, Xu, and Zha]{li2024fouriermamba}
Dong Li, Yidi Liu, Xueyang Fu, Senyan Xu, and Zheng-Jun Zha.
\newblock Fouriermamba: Fourier learning integration with state space models for image deraining.
\newblock \emph{arXiv preprint arXiv:2405.19450}, 2024.

\bibitem[Li et~al.(2018)Li, Xie, Zhao, Wei, Gu, Tao, and Meng]{Li2018VideoRainStreak}
Minghan Li, Qi Xie, Qian Zhao, Wei Wei, Shuhang Gu, Jing Tao, and Deyu Meng.
\newblock Video rain streak removal by multiscale convolutional sparse coding.
\newblock In \emph{CVPR}, 2018.

\bibitem[Lieber et~al.(2024)Lieber, Lenz, Bata, Cohen, Osin, Dalmedigos, Safahi, Meirom, Belinkov, Shalev-Shwartz, et~al.]{lieber2024jamba}
Opher Lieber, Barak Lenz, Hofit Bata, Gal Cohen, Jhonathan Osin, Itay Dalmedigos, Erez Safahi, Shaked Meirom, Yonatan Belinkov, Shai Shalev-Shwartz, et~al.
\newblock Jamba: A hybrid transformer-mamba language model.
\newblock \emph{arXiv preprint arXiv:2403.19887}, 2024.

\bibitem[Liu et~al.(2018)Liu, Yang, Yang, and Guo]{Liu2018Erase}
Jiaying Liu, Wenhan Yang, Shuai Yang, and Zongming Guo.
\newblock Erase or fill? deep joint recurrent rain removal and reconstruction in videos.
\newblock In \emph{CVPR}, 2018.

\bibitem[Liu et~al.(2019)Liu, Yang, Yang, and Guo]{Liu2019D3RNet}
Jiaying Liu, Wenhan Yang, Shuai Yang, and Zongming Guo.
\newblock D3r-net: Dynamic routing residue recurrent network for video rain removal.
\newblock \emph{IEEE TIP}, 2019.

\bibitem[Luo et~al.(2015)Luo, Xu, and Ji]{luo2015removing}
Yu Luo, Yong Xu, and Hui Ji.
\newblock Removing rain from a single image via discriminative sparse coding.
\newblock In \emph{ICCV}, 2015.

\bibitem[Redmon(2018)]{redmon2018yolov3}
Joseph Redmon.
\newblock Yolov3: An incremental improvement.
\newblock \emph{arXiv preprint arXiv:1804.02767}, 2018.

\bibitem[Ren et~al.(2017)Ren, Tian, Han, Chan, and Tang]{Ren2017VideoDesnowing}
Weihong Ren, Jiandong Tian, Zhi Han, Antoni Chan, and Yandong Tang.
\newblock Video desnowing and deraining based on matrix decomposition.
\newblock In \emph{CVPR}, 2017.

\bibitem[Ronneberger et~al.(2015)Ronneberger, Fischer, and Brox]{Ronneberger2015UNetCN}
Olaf Ronneberger, Philipp Fischer, and Thomas Brox.
\newblock U-net: Convolutional networks for biomedical image segmentation.
\newblock In \emph{MICCAI}, 2015.

\bibitem[Santhaseelan and Asari(2015)]{Santhaseelan2015Utilizing}
Varun Santhaseelan and Vijayan~K. Asari.
\newblock Utilizing local phase information to remove rain from video.
\newblock \emph{IJCV}, 2015.

\bibitem[Sindagi et~al.(2020)Sindagi, Oza, Yasarla, and Patel]{sindagi2020prior}
Vishwanath~A Sindagi, Poojan Oza, Rajeev Yasarla, and Vishal~M Patel.
\newblock Prior-based domain adaptive object detection for hazy and rainy conditions.
\newblock In \emph{ECCV}, 2020.

\bibitem[Smith et~al.(2022)Smith, Warrington, and Linderman]{smith2022simplified}
Jimmy~TH Smith, Andrew Warrington, and Scott~W Linderman.
\newblock Simplified state space layers for sequence modeling.
\newblock \emph{arXiv preprint arXiv:2208.04933}, 2022.

\bibitem[Sun et~al.(2022)Sun, Ren, Wang, and Cao]{sun2022rethinking}
Shangquan Sun, Wenqi Ren, Tao Wang, and Xiaochun Cao.
\newblock Rethinking image restoration for object detection.
\newblock \emph{NeurIPS}, 2022.

\bibitem[Sun et~al.(2023)Sun, Ren, Li, Zhang, Liang, and Cao]{sun2023event}
Shangquan Sun, Wenqi Ren, Jingzhi Li, Kaihao Zhang, Meiyu Liang, and Xiaochun Cao.
\newblock Event-aware video deraining via multi-patch progressive learning.
\newblock \emph{IEEE TIP}, 2023.

\bibitem[Sun et~al.(2024{\natexlab{a}})Sun, Ren, Gao, Wang, and Cao]{sun2025restoring}
Shangquan Sun, Wenqi Ren, Xinwei Gao, Rui Wang, and Xiaochun Cao.
\newblock Restoring images in adverse weather conditions via histogram transformer.
\newblock In \emph{ECCV}, 2024{\natexlab{a}}.

\bibitem[Sun et~al.(2024{\natexlab{b}})Sun, Ren, Li, Wang, and Cao]{sun2024logit}
Shangquan Sun, Wenqi Ren, Jingzhi Li, Rui Wang, and Xiaochun Cao.
\newblock Logit standardization in knowledge distillation.
\newblock In \emph{Proceedings of the IEEE/CVF conference on computer vision and pattern recognition}, pages 15731--15740, 2024{\natexlab{b}}.

\bibitem[Sun et~al.(2024{\natexlab{c}})Sun, Ren, Zhou, Gan, Wang, and Cao]{sun2024hybrid}
Shangquan Sun, Wenqi Ren, Juxiang Zhou, Jianhou Gan, Rui Wang, and Xiaochun Cao.
\newblock A hybrid transformer-mamba network for single image deraining.
\newblock \emph{arXiv preprint arXiv:2409.00410}, 2024{\natexlab{c}}.

\bibitem[Wang et~al.(2018)Wang, Xu, Wang, and Tao]{wang2018perceptual}
Chaoyue Wang, Chang Xu, Chaohui Wang, and Dacheng Tao.
\newblock Perceptual adversarial networks for image-to-image transformation.
\newblock \emph{IEEE TIP}, 2018.

\bibitem[Wang et~al.(2020)Wang, Xie, Zhao, and Meng]{wang2020modeldriven}
Hong Wang, Qi Xie, Qian Zhao, and Deyu Meng.
\newblock A model-driven deep neural network for single image rain removal.
\newblock In \emph{CVPR}, 2020.

\bibitem[Wei et~al.(2017)Wei, Yi, Xie, Zhao, Meng, and Xu]{Wei2017Should}
Wei Wei, Lixuan Yi, Qi Xie, Qian Zhao, Deyu Meng, and Zongben Xu.
\newblock Should we encode rain streaks in video as deterministic or stochastic?
\newblock In \emph{ICCV}, 2017.

\bibitem[Wei et~al.(2019)Wei, Meng, Zhao, Xu, and Wu]{wei2019semi}
Wei Wei, Deyu Meng, Qian Zhao, Zongben Xu, and Ying Wu.
\newblock Semi-supervised transfer learning for image rain removal.
\newblock In \emph{CVPR}, 2019.

\bibitem[Wu et~al.(2024)Wu, Yang, Xu, Wang, Zhou, and Zhu]{wu2024rainmamba}
Hongtao Wu, Yijun Yang, Huihui Xu, Weiming Wang, Jinni Zhou, and Lei Zhu.
\newblock Rainmamba: Enhanced locality learning with state space models for video deraining.
\newblock In \emph{ACM MM}, 2024.

\bibitem[Xie et~al.(2024)Xie, Cui, Tan, Zheng, and Yu]{xie2024fusionmamba}
Xinyu Xie, Yawen Cui, Tao Tan, Xubin Zheng, and Zitong Yu.
\newblock Fusionmamba: Dynamic feature enhancement for multimodal image fusion with mamba.
\newblock \emph{Visual Intelligence}, 2\penalty0 (1):\penalty0 37, 2024.

\bibitem[Yamashita and Ikehara(2024)]{yamashita2024image}
Shugo Yamashita and Masaaki Ikehara.
\newblock Image deraining with frequency-enhanced state space model.
\newblock \emph{arXiv preprint arXiv:2405.16470}, 2024.

\bibitem[Yan et~al.(2021)Yan, Tan, Yang, and Dai]{Yan_2021_CVPR}
Wending Yan, Robby~T. Tan, Wenhan Yang, and Dengxin Dai.
\newblock Self-aligned video deraining with transmission-depth consistency.
\newblock In \emph{CVPR}, 2021.

\bibitem[Yang et~al.(2019)Yang, Liu, and Feng]{Yang2019Frame}
Wenhan Yang, Jiaying Liu, and Jiashi Feng.
\newblock Frame-consistent recurrent video deraining with dual-level flow.
\newblock In \emph{CVPR}, 2019.

\bibitem[Yang et~al.(2020{\natexlab{a}})Yang, Tan, Feng, Guo, Yan, and Liu]{yang2020joint}
Wenhan Yang, Robby~T. Tan, Jiashi Feng, Zongming Guo, Shuicheng Yan, and Jiaying Liu.
\newblock Joint rain detection and removal from a single image with contextualized deep networks.
\newblock \emph{IEEE TPAMI}, 2020{\natexlab{a}}.

\bibitem[Yang et~al.(2020{\natexlab{b}})Yang, Tan, Wang, and Liu]{Yang2020Self}
Wenhan Yang, Robby~T. Tan, Shiqi Wang, and Jiaying Liu.
\newblock Self-learning video rain streak removal: When cyclic consistency meets temporal correspondence.
\newblock In \emph{CVPR}, 2020{\natexlab{b}}.

\bibitem[Yang et~al.(2022{\natexlab{a}})Yang, Tan, Feng, Wang, Cheng, and Liu]{Yang2022MFGAN}
Wenhan Yang, Robby~T. Tan, Jiashi Feng, Shiqi Wang, Bin Cheng, and Jiaying Liu.
\newblock Recurrent multi-frame deraining: Combining physics guidance and adversarial learning.
\newblock \emph{IEEE TPAMI}, 2022{\natexlab{a}}.

\bibitem[Yang et~al.(2022{\natexlab{b}})Yang, Tan, Wang, Kot, and Liu]{yang2022learning}
Wenhan Yang, Robby~T. Tan, Shiqi Wang, Alex~C. Kot, and Jiaying Liu.
\newblock Learning to remove rain in video with self-supervision.
\newblock \emph{IEEE TPAMI}, 2022{\natexlab{b}}.

\bibitem[Yasarla and Patel(2019)]{Yasarla2019Uncertaity}
Rajeev Yasarla and Vishal~M. Patel.
\newblock Uncertainty guided multi-scale residual learning-using a cycle spinning cnn for single image de-raining.
\newblock In \emph{CVPR}, 2019.

\bibitem[You et~al.(2016)You, Tan, Kawakami, Mukaigawa, and Ikeuchi]{You2016Adherent}
Shaodi You, Robby~T. Tan, Rei Kawakami, Yasuhiro Mukaigawa, and Katsushi Ikeuchi.
\newblock Adherent raindrop modeling, detection and removal in video.
\newblock \emph{IEEE TPAMI}, 2016.

\bibitem[Yu and Wang(2024)]{yu2024mambaout}
Weihao Yu and Xinchao Wang.
\newblock Mambaout: Do we really need mamba for vision?
\newblock \emph{arXiv preprint arXiv:2405.07992}, 2024.

\bibitem[Yue et~al.(2021)Yue, Xie, Zhao, and Meng]{Yue2021Semi}
Zongsheng Yue, Jianwen Xie, Qian Zhao, and Deyu Meng.
\newblock Semi-supervised video deraining with dynamical rain generator.
\newblock In \emph{CVPR}, 2021.

\bibitem[Zamir et~al.(2022)Zamir, Arora, Khan, Hayat, Khan, and Yang]{zamir2022restormer}
Syed~Waqas Zamir, Aditya Arora, Salman Khan, Munawar Hayat, Fahad~Shahbaz Khan, and Ming–Hsuan Yang.
\newblock Restormer: Efficient transformer for high-resolution image restoration.
\newblock In \emph{CVPR}, 2022.

\bibitem[Zhang et~al.(2020)Zhang, Sindagi, and Patel]{zhang2020image}
He Zhang, Vishwanath Sindagi, and Vishal~M. Patel.
\newblock Image de-raining using a conditional generative adversarial network.
\newblock \emph{IEEE TCSVT}, 2020.

\bibitem[Zhang et~al.(2022)Zhang, Li, Luo, Ren, and Liu]{zhang2022enhanced}
Kaihao Zhang, Dongxu Li, Wenhan Luo, Wenqi Ren, and Wei Liu.
\newblock Enhanced spatio-temporal interaction learning for video deraining: A faster and better framework.
\newblock \emph{IEEE TPAMI}, 2022.

\bibitem[Zhang et~al.(2006)Zhang, Li, Qi, Leow, and Ng]{zhang2006rain}
Xiaopeng Zhang, Hao Li, Yingyi Qi, Wee~Kheng Leow, and Teck~Khim Ng.
\newblock Rain removal in video by combining temporal and chromatic properties.
\newblock In \emph{ICME}, 2006.

\bibitem[Zhen et~al.(2024)Zhen, Hu, and Feng]{zhen2024freqmamba}
Zou Zhen, Yu Hu, and Zhao Feng.
\newblock Freqmamba: Viewing mamba from a frequency perspective for image deraining.
\newblock \emph{arXiv preprint arXiv:2404.09476}, 2024.

\bibitem[Zhou et~al.(2022)Zhou, Yin, Koltun, and Krähenbühl]{zhou2022gtr}
Xingyi Zhou, Tianwei Yin, Vladlen Koltun, and Philipp Krähenbühl.
\newblock Global tracking transformers.
\newblock In \emph{CVPR}, 2022.

\bibitem[Zhu et~al.(2024)Zhu, Liao, Zhang, Wang, Liu, and Wang]{zhu2024visionmamba}
Lianghui Zhu, Bencheng Liao, Qian Zhang, Xinlong Wang, Wenyu Liu, and Xinggang Wang.
\newblock Vision mamba: Efficient visual representation learning with bidirectional state space model.
\newblock \emph{arXiv preprint arXiv:2401.09417}, 2024.

\bibitem[Zou et~al.(2024)Zou, Yu, Huang, and Zhao]{zou2024freqmamba}
Zhen Zou, Hu Yu, Jie Huang, and Feng Zhao.
\newblock Freqmamba: Viewing mamba from a frequency perspective for image deraining.
\newblock In \emph{ACM MM}, 2024.

\end{thebibliography}
}


\end{document}